\documentclass[]{interact}

\usepackage{epstopdf}

\usepackage[utf8]{inputenc} 
\usepackage[T1]{fontenc}    
\usepackage{url}            
\usepackage{booktabs}       
\usepackage{amsfonts}       
\usepackage{nicefrac}       

\usepackage{mathtools}

\usepackage{algorithm}
\usepackage{algpseudocode}
\usepackage{graphicx}
\usepackage{wrapfig}
\usepackage{float}
\usepackage{MnSymbol}%
\usepackage{wasysym}

\usepackage{microtype}      
\usepackage{xcolor}         
\usepackage{graphicx}
\usepackage{subcaption}

\usepackage{amsmath}
\usepackage{multicol}
\usepackage{multirow}
\usepackage{tabularx}
\usepackage{placeins}
\usepackage{adjustbox}
\usepackage[colorlinks = true,
            linkcolor = blue,
            urlcolor  = blue,
            citecolor = blue]{hyperref}

\def\*#1{\mathbf{#1}}
\def\+#1{\mathcal{#1}}
\def\emp*#1{{#1}_n}

\newcommand*{\eqdef}{\vcentcolon =}

\definecolor{blush}{rgb}{0.87, 0.36, 0.51}

\usepackage[some]{background}
\backgroundsetup{%
scale=1,  
angle=0,  
opacity=0.7,  
color = {darkgray},  
contents={\parbox{\textwidth}{\centering 
\large This paper has been accepted for publication in \\ International Journal of Geographical Information Science (IJGIS) \\[26cm]}
}
}

\usepackage{natbib}
\bibpunct[, ]{(}{)}{,}{a}{}{,}
\renewcommand\bibfont{\fontsize{10}{12}\selectfont}

\theoremstyle{plain}

\theoremstyle{definition}

\theoremstyle{remark}

\begin{document}


\title{GeOT: A spatially explicit framework for evaluating spatio-temporal predictions}

\author{
\name{Nina Wiedemann\textsuperscript{a}\thanks{CONTACT N.~W. e-mail: nwiedemann@ethz.ch}, Théo Uscidda\textsuperscript{b}, Martin Raubal\textsuperscript{a}}
\affil{\textsuperscript{a}Chair of Geoinformation Engineering, ETH Zürich\; \textsuperscript{b}CREST-ENSAE, Paris, France}
}

\maketitle

\BgThispage{}

\begin{abstract}
When predicting observations across space and time, the spatial layout of errors impacts a model's real-world utility. For instance, in bike sharing demand prediction, error patterns translate to relocation costs. However, commonly used error metrics in GeoAI evaluate predictions point-wise, neglecting effects such as spatial heterogeneity, autocorrelation, and the Modifiable Areal Unit Problem. 
We put forward Optimal Transport (OT) as a spatial evaluation metric and loss function. 
The proposed framework, called GeOT, assesses the performance of prediction models by quantifying the transport costs associated with their prediction errors. 
Through experiments on real and synthetic data, we demonstrate that 1) the spatial distribution of prediction errors relates to real-world costs in many applications, 2) OT captures these spatial costs more accurately than existing metrics, and 3) OT enhances comparability across spatial and temporal scales. 
Finally, we advocate for leveraging OT as a loss function in neural networks to improve the spatial accuracy of predictions. Experiments with bike sharing, charging station, and traffic datasets show that spatial costs are significantly reduced with only marginal changes to non-spatial error metrics. Thus, this approach not only offers a spatially explicit tool for model evaluation and selection, but also integrates spatial considerations into model training.
\end{abstract}


\section{Introduction}

Geographic Information Science (GIScience) aims to develop analysis and prediction tools tailored to the specific challenges involved with geographic data, true to the principle that ``spatial is special''. Meanwhile, the vast majority of GeoAI research aims to implement a spatially-explicit \textit{model design}~\citep{janowicz2020geoai,hu2019geoai, liu2022review}, while spatial considerations in the model \textit{evaluation} are neglected. 
Consider the following examples: In weather forecasting, locating a rain shower 50km from its actual occurrence is clearly worse than mislocating it by 5km. Local deviations in traffic forecasts are less severe than an
occurrence of traffic congestion far from the expected location. Errors in predicting wildfire spread involve costly re-allocation of firefighting resources, which are the more time-consuming the farther the predicted direction of fire spread is from its real direction. In sum, the \textit{spatial distribution} of prediction errors plays an important role for many applications in GIScience and transportation, and evaluation frameworks should account for the costs arising from relocation or resource allocation effort.

Currently, GeoAI methods are evaluated with standard error metrics such as the mean squared error (MSE) or mean absolute percentage error (MAPE). These metrics average the error over locations, ignoring the impact of prediction errors on downstream tasks such as bike relocation, traffic management, or mission planning for firefighting. Consequently, there is a mismatch between how predictions are evaluated and their practical utility~\citep{yan2022integrating}. This can be addressed through simulations that evaluate the real-world impact of prediction errors~\citep{peled2021quality}; however such simulations are highly application-specific and cumbersome to create. 
On the other hand, spatial statistics offers application-independent measures of the errors' spatial distribution, for example using residual autocorrelation to evaluate a model's ability to account for spatial heterogeneity~\citep{zhang2009evaluation, chen2016spatial}. However, these methods cannot measure relocation or resource allocation costs and lack interpretability and robustness towards their parameters~\citep{chou1993critical}.

\begin{figure}[htb]
    \centering
    \includegraphics[width=\textwidth]{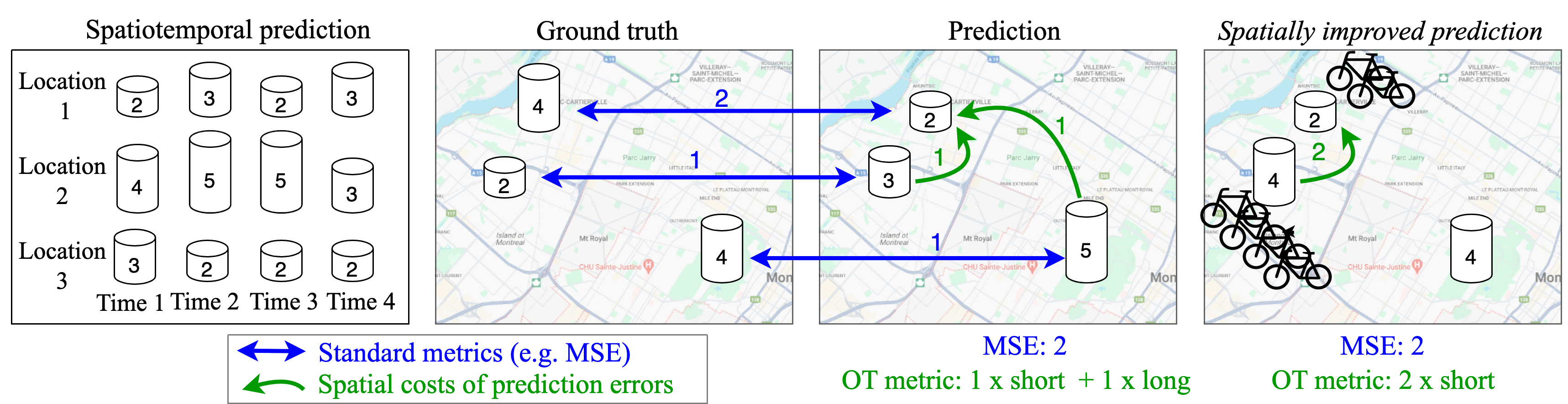}
    \caption{Optimal transport as an evaluation framework in geospatial data science. Spatio-temporal prediction problems involve forecasting spatial observations over time, such as estimating bike-sharing demand at multiple stations (left). Conventional metrics usually treat locations independently, ignoring their spatial distribution (blue). In contrast, our GeOT framework based on Optimal Transport accounts for spatial costs, quantifying prediction errors in terms of the effort required for relocation or resource allocation, such as relocating bicycles between stations (green).}
    \label{fig:motivation}
\end{figure}

In this work, we propose to evaluate spatial prediction models with Optimal Transport (OT). OT is a mathematical framework providing methods to measure the disparity between two distributions. 
Here, we show how OT can be leveraged to compare the real and predicted spatial distributions in geospatial applications. As illustrated in \autoref{fig:motivation}, the proposed OT metric - named GeOT - measures the \textit{spatial costs} of prediction errors in terms of the redistribution effort necessary to align the predictions with the ground truth. 
Crucially, OT is not limited to Euclidean distances, but can reflect application-specific and interpretable costs such as user relocation time or monetary costs for resource reallocation. Our framework is based on partial OT~\citep{guittet2002extended, piccoli2014generalized, maas2015generalized} and is applicable to diverse spatio-temporal prediction tasks 
(see \autoref{tab:ot_applications}). Our contribution is twofold: First, we introduce OT as an evaluation metric, highlighting its relevance for GeoAI in two case studies and analyzing its relationship to existing measures. Secondly, we demonstrate how this evaluation metric naturally informs the design of a new loss function for GeoAI models. By directly minimizing the OT error during training, this loss function has the potential to capture spatial dependencies and application-specific costs more effectively than traditional loss functions. 

The remainder of the paper is structured as follows: In \autoref{sec:methods}, we introduce OT-based metrics for spatio-temporal predictions, positioning our work within the broader context of OT and GeoAI research. In \autoref{sec:case_study} and \autoref{sec:interpolation} we empirically illustrate the potential of OT as an \textit{evaluation} framework in GIScience, and \autoref{sec:outlook} presents experiments using this novel metric as a \textit{loss} function. Finally, we discuss challenges and limitations of applying OT in \autoref{sec:discussion} and present conclusions in \autoref{sec:conclusion}.

\begin{table}[ht] 
    \centering 
    \setlength{\arrayrulewidth}{0.5mm}  
    \setlength{\tabcolsep}{5pt}  
    \renewcommand{\arraystretch}{1.3}  

    \resizebox{1.0\textwidth}{!}{
    \begin{tabularx}{1.5\textwidth}{|X|X|X|X|X|X|}
        \hline
        \textbf{Application} & \textbf{Category} & \textbf{Relevance} & \textbf{Location} & \textbf{Spatial Costs} & \textbf{Interpretation} \\ \hline

        Wildfire spread prediction & Vision-based GeoAI & Improve resource allocation for firefighting & Cells in geographic raster data & Operational costs for relocating resources between cells & Resource reallocation costs (distance between predicted and true fire spread) \\ \hline

        Weather forecasting & Vision-based GeoAI & Adequate preparation and planning & Cells in geographic raster data & Euclidean distance & OT error indicates spatial displacement of an occurring weather phenomenon with respect to the prediction \\ \hline

        Bike, scooter, or car sharing demand prediction & Spatiotemporal time series forecasting & Efficient allocation of supply & Stations of shared system & Map-matched driving distances (relocation costs) & Costs for resource reallocations or for users to relocate due to prediction errors \\ \hline

        EV charging station occupancy prediction & Spatiotemporal time series forecasting & Supporting navigation & EV charging stations & Map-matched driving distance & Opportunity costs for drivers to relocate due to prediction errors \\ \hline

        Deforestation rate estimation & Spatial interpolation & Inform policy and intervention measures by identifying areas at high risk of deforestation & Towns or regions & Communication distance between places & OT error indicates spatial mismatch between potential intervention measures and actual deforestation \\ \hline

        Estimation of heavy metal pollution & Spatial interpolation & Monitoring and intervention against pollution & Measurement locations & Euclidean distance & Spatial displacement of pollution estimates and opportunity costs \\ \hline
    \end{tabularx} }

    \caption{Potential applications for spatial evaluation metrics, such as the GeOT framework. Spatial costs are defined between locations in the form of raster cells, points, or regions. The user-defined cost matrix allows for an application-specific interpretation of the spatial prediction error, e.g., as relocation costs, resource allocation, or opportunity costs.}
    \label{tab:ot_applications}
\end{table}


\section{Methods: Optimal Transport metrics for spatio-temporal predictions}\label{sec:methods}

\subsection{The Optimal Transport framework}
\label{subsec:ot-framework}

Optimal transport (OT) is a mathematical framework for comparing probability distributions \citep{santambrogio2015optimal} and has recently become increasingly influential in the field of machine learning \citep{peyre2019computational}. This growing interest has led to significant methodological advancements for computing OT, especially in high-dimensional and continuous settings. Notable developments include the use of convex neural networks~\citep{makkuva2020optimal,korotin2021continuous,huang2021convex} or the \textit{Monge} gap~\citep{uscidda2023monge} to efficiently approximate OT solutions, normalizing flows~\citep{tong2020trajectorynet,lipman2023flow,pooladian2023multisample,tong2023conditional,tong2023simulation,klein2024entropicgromovwassersteinflow,eyring2024unbalancednessneuralmongemaps}, and the integration of OT for disentangled representation learning~\citep{nakagawa2023gromovwassersteinautoencoders,uscidda2024disentangledrepresentationlearninggeometry}.

Solving an OT problem involves finding the most cost-effective way to transport a source distribution $\mu$ to a target distribution $\nu$. Focusing here on discrete distributions, let $\mu = \sum_{i=1}^n \*p_i \delta_{\*x_i}$ and $\nu = \sum_{i=1}^m \*q_j \delta_{\*y_j}$, where $\*p = (\*p_1, \ldots, \*p_n)$ and $\*q = (\*q_1, \ldots, \*q_m)$ are histograms and $\*x_1, \ldots, \*x_n$ and $\*y_1, \ldots, \*y_m$ are the locations in $\mathbb{R}^d$ where the mass of each measure lies. Additionally, let $c : \mathbb{R}^d \times \mathbb{R}^d \to \mathbb{R}$ be a cost function, s.t.\ $c(\*x, \*y)$ measures the cost of moving a unit of mass from location $\*x$ to location $\*y$. $\*C \eqdef [c(\*x_i, \*y_j)]_{1 \leq i,j \leq n,m} \in \mathbb{R}^{n\times m}$ is called the cost matrix. The goal of OT is to transport $\mu$ onto $\nu$ through a coupling matrix $\*T \in \mathrm{U}(\*p, \*q) \eqdef \{\*T \in \mathbb{R}_+^{n \times n} \mid \*T \mathrm{1}_n = \*p, \, \*T^\top \mathrm{1}_m = \*q\}$ while minimizing the cost of transportation quantified by $c$. Here, $\*T_{ij}$ denotes the amount of mass transported from $\*x_i$ to $\*y_j$. In sum, OT aims to solve the following optimization problem:
\begin{equation}
\label{eq:discrete-ot}
\min_{\*T\in \mathrm{U}(\*p, \*q)} \sum_{i,j=1}^{n,m}\*T_{ij}\*C_{ij} \,\, \Leftrightarrow \,\, \min_{\*T\in\mathrm{U}(\*p, \*q)} \langle \*T, \*C \rangle
\end{equation}
where $\langle \cdot, \cdot \rangle$ denotes the Frobenius inner product. The OT problem~\eqref{eq:discrete-ot} is a linear program, which can be solved using, e.g., the network-simplex algorithm~\citep{bertsimas-LPbook}. A solution $\*T^\star$ to this problem, which always exists, is called an OT coupling. The $c$-Wasserstein distance is defined as: 
\begin{equation}
\label{eq:wasserstein-distance}
\mathrm{W}_c(\mu,\nu) = \min_{\*T\in\mathrm{U}(\*p, \*q)} \langle \*T, \*C \rangle\, = \sum_{i,j=1}^{n,m}\*T^\star_{ij}\*C_{ij}.
\end{equation}
This quantity is also referred to as the Earth Mover's Distance (EMD)~\citep{yubner2000earth}. When $c(\*x, \*y) = \|\*x - \*y\|_2^q$ for any $p \geq 1$, $\mathrm{W}_{c}(\mu, \nu) = 0$ i.f.f.\ $\mu = \nu$~\citep[Prop.\ 5.1]{santambrogio2015optimal}. As a result, $\mathrm{W}_{c}$ provides a natural quantity to compare distributions. 

\subsection{Optimal Transport for evaluating spatio-temporal predictions}\label{sec:methods_eval}

In this work, we investigate the use of this distance as an evaluation metric and loss function for geospatial prediction problems. Assume that a GeoAI model provides predictions at fixed locations, such as pickups at bike sharing stations (see \autoref{fig:motivation}).  To repurpose OT for spatial evaluation, we set $\mu$ to the predicted and $\nu$ to the true spatial distribution of observations. This approach utilizes the Wasserstein distance between \textit{signatures}, a special case where both distributions are defined over the same locations ($n=m$ and $\*x_i = \*y_i$, $\forall i\in \{1,\dots,n\}$), but with different histograms. Let $\*x_1, \dots \*x_n$ be the spatial locations, and let $\*o = (\*o_1, \dots, \*o_n)$ be the true observations and $\hat{\*o} = (\hat{\*o}_1, \dots, \hat{\*o}_n)$ the predicted observations at these locations. Thus, the source and target distribution ($\mu$ and $\nu$) are set to the predicted spatial observations, $\mu=\sum_{i=1}^n \hat{\*o}_i \delta_{\mathbf{x}_i} $, and the true observations $\nu=\sum_{i=1}^n \*o_i \delta_{\mathbf{x}_i} $, respectively. For instance, $\*o_i$ could represent the demand for shared bicycles at the $i$-th bike sharing station, located at $\*x_i$. For geospatial data, the locations $\*x$ are usually two-dimensional, $\*x\in \mathbb{R}^2$.

By selecting an appropriate cost function \( c \) between locations and setting \( \mathbf{C}_{ij} = c(\mathbf{x}_i, \mathbf{x}_j) \), we can solve Problem~\eqref{eq:discrete-ot} between \( \mu \) and \( \nu \) to compute \( \mathrm{W}_c(\mu, \nu) \). 
In sum, we define the cost-dependent geospatial OT error as
\begin{align}\label{eq:oterror}
    \mathrm{W}^{geo}_c = \mathrm{W}_c(\mu, \nu) \text{ with } \mu=\sum_{i=1}^n \hat{\*o}_i \delta_{\mathbf{x}_i},\  \nu=\sum_{i=1}^n \*o_i \delta_{\mathbf{x}_i}
\end{align}
This value translates to the minimal cost necessary to align the predicted with the true spatial distribution of observations. In other words, $\mathrm{W}^{geo}_c(\mu, \nu)$ measures the total spatial costs to ``undo'' errors of the predictive model. 

\begin{figure}
    \centering
    \includegraphics[width=\textwidth]{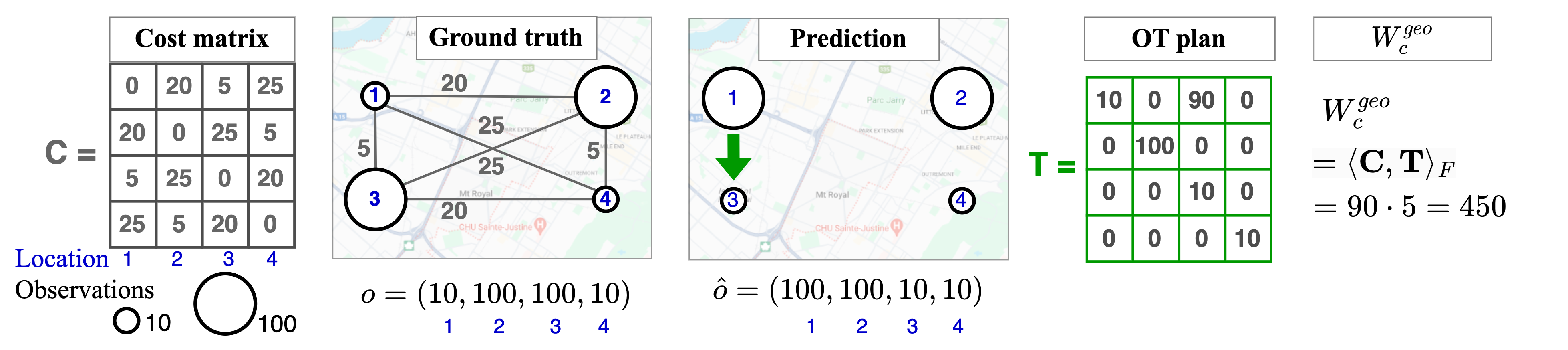}
    \caption{Quantifying spatial costs with Optimal Transport. Given a cost matrix $C$ defined between location pairs, prediction errors are measured in terms of the minimal transport costs required to align the predictions with the true observations (see \autoref{sec:methods_eval}). In the example, a mass of 90 must be transported from location 1 to location 3 with cost 5, leading to an OT error of 450.}
    \label{fig:notation}
\end{figure}

\autoref{fig:notation} provides an example. 
The prediction error, here an overestimation of $\*o_1$ and underestimation of $\*o_3$, results in transportation costs of 90 between location 1 and 3 ($\*T_{13}=90$). The total costs are thus $\mathrm{W}_c^{geo} = \sum_{i,j=1}^{n,m}\*T^\star_{ij}\*C_{ij} =  90 \cdot \*C_{13} = 450$. 
In general, suitable cost functions include, but are not limited to any \( p \)-norm–distance between locations, map-matched distances, monetary costs or CO\(_2\) emissions. If the cost matrix corresponds to the (squared) Euclidean distance between the geographic locations, i.e., $\*C_{ij} = c(\*x_i, \*x_j) = \|\*x_i - \*x_j\|_2^2$, this corresponds to the 2-Wasserstein distance, $\mathrm{W}^2_2(\mu, \nu)$. 

While we have focused on spatial predictions with \textit{fixed} locations ($\*x = \*y$), it is important to note that the metric can be easily extended to applications where the predicted locations differ from the locations of the true observations ($n\ne m$ and $\*x_i \ne \*y_i$). An example of this is provided in \autoref{sec:unpairedot}, where we predict the locations of high mortality rates and measure the spatial error of these predictions using OT.

\subsection{Partial Optimal Transport}\label{sec:partialot}

The standard OT formulation assumes equal total mass in both distributions, which is unrealistic in our case without normalization. Unbalanced OT, introduced by \citet{kantorovich1958space}, relaxes this constraint by minimizing the divergence between the distributions and the transportation plan marginals~\citep{chapel2021unbalanced, chizat2018scaling, liero2018optimal, sejourne2019sinkhorn, pham2020unbalanced}. However, this approach lacks interpretability in geospatial applications, as the transportation matrix is not executable in practice and the relaxation blurs the distinction between costs from distribution mismatch and overall mass imbalance. Partial OT, in contrast, addresses mass imbalance by assigning explicit costs to untransported mass using methods like "dummy points"\citep{chapel2020partial}, "dustbin"\citep{sarlin2020superglue}, or "waste vectors"~\citep{guittet2002extended}. While better suited to our needs, its general formulation allows a large fraction of the mass to vanish. Instead, we aim to transport as much mass as possible to balance out all prediction errors. This aligns with a special case in \citet{chapel2020partial}, who show that the problem reduces to standard OT when extending the measures and cost matrix.

Following \citet{chapel2020partial}, we add a dummy location $\*x_{n+1}$ in both source and target measures. The mass at this dummy point is set to zero or the total mass difference ($|\sum_{i=1}^n \*o_i - \sum_{i=1}^n \hat{\*o}_i|$) respectively, dependent on whether the source or target distribution has larger mass. Formally, let $s = \min(\sum_{i=1}^n \*o_i, \sum_{i=1}^n \hat{\*o}_i)$, and we define $\*o_{n+1} = \sum_{i=0}^n \hat{\*o}_i - s$, and $\hat{\*o}_{n+1} = \sum_{i=0}^n \*o_i - s$. For example, if the sum of observations over all locations is 10 ($\sum_{i=1}^n \*o_i = 10$), and the predicted total is 12, we add a dummy location with $\*o_{n+1} = 2$ and $\hat{\*o}_{n+1}=0$. We denote the adapted measures including the dummy points as $\tilde{\mu}$ and $\tilde{\nu}$, which now have equal mass by design. The cost matrix is adapted to penalize the overshooting mass with a fixed cost $\phi$: 
\begin{gather*}
    \widetilde{\*C}(\phi) = 
\left(
{\begin{array}{cccc}
c_{11} & \dots & c_{1n} & \phi \\
\dots & \ddots & \dots & \dots \\
c_{n1} & \dots & c_{nn} & \phi \\
\phi & \dots & \phi & \phi \\
\end{array}
}\right)\\
\end{gather*}
As \citet{chapel2020partial} show, partial OT corresponds to solving balanced OT on $\tilde{\mu}, \tilde{\nu}$ and $\widetilde{\*C}$. Thus, we define: 
\begin{align}
    \mathrm{W}^{geo}_{c, \phi} = \mathrm{W}_{\tilde{c}}(\tilde{\mu}, \tilde{\nu}) \text{ with } \tilde{\mu}=\sum_{i=1}^{n+1} \hat{\*o}_i \delta_{\mathbf{x}_i},\  \tilde{\nu}=\sum_{i=1}^{n+1} \*o_i \delta_{\mathbf{x}_i}
\end{align}

The solution yields a transportation matrix that contains the flow of mass between locations, as well as the outflow or inflow dependent on the total mass difference. 
In our evaluation framework, $\mathrm{W}^{geo}_{c, \phi}$ combines the total prediction error and the spatial distributional error, with $\phi$ controlling their emphasis: Higher $\phi$ puts more weight on the total error $\sum_{i=1}^n \*o_i - \sum_{i=1}^n \hat{\*o}_i$, while lower $\phi$ highlights the distributional error. It is worth noting that the penalty $\phi$ could easily be defined in a location-dependent manner; i.e., penalizing the import / export to some locations more than to others. For instance, this could be useful when considering predictions of bike sharing demand, where bikes are transported from a distribution center to the stations.

\subsection{OT-based loss function based on Sinkhorn divergences}\label{sec:sinkhorn}

A natural progression for the OT error is its integration into the \textit{training} of neural networks as a spatial loss function. However, $\mathrm{W}_c$ is non-differentiable with respect to its inputs, impeding its direct use as a loss function. One way to alleviate these challenges is to rely on entropic regularization~\citep{cuturi_sinkhorn_2013}. Introducing $H(\*T) = \sum_{i,j=1}^{n,m}\*T_{ij}\log(\*T_{ij})$ and $\varepsilon > 0$, the Entropic OT problem between $\mu = \sum_{i=1}^n\*p_i\delta_{\*x_i}$ and $\nu = \sum_{i=1}^m\*q_i\delta_{\*y_i}$ is defined as
\begin{equation}
\label{eq:entropic-ot}
    \mathrm{W}_{c,\varepsilon}(\mu,\nu) = \min_{\*T\in\mathrm{U}(\*p, \*q)} \langle \*T, \*C \rangle\, - \varepsilon H(\*T).
\end{equation}
\citeauthor{Sinkhorn64}'s algorithm provides an iterative approach for finding a unique solution to the dual formulation of (\ref{eq:entropic-ot}). By \citeauthor{danskin2012theory}'s Theorem, the uniqueness of the solution guarantees the differentiability of $\mathrm{W}_{c,\varepsilon}(\mu,\nu)$ with respect to its inputs, allowing its use as a loss function. To correct biases in this loss function it was proposed to center the Entropic OT objective~\citep{genevay2018learning,feydy2018interpolating,pooladian2022debiaserbewarepitfallscentering}, defining the \textit{Sinkhorn divergence} as follows:
\begin{equation}
    \mathrm{S}_{c, \varepsilon}(\mu, \nu) = \mathrm{W}_{c,\varepsilon}(\mu, \nu) - \tfrac{1}{2}(\mathrm{W}_{c,\varepsilon}(\mu, \mu) + \mathrm{W}_{c,\varepsilon}(\nu, \nu))
\end{equation}
A more comprehensive explanation of the Sinkhorn divergence is given in \autoref{app:sinkhorn}. In practice, we use the implementation provided in the \texttt{geomloss} package~\citep{feydy2018interpolating}, which employs the Sinkhorn algorithm. 

\subsection{Related work}

While OT has become a popular tool in other applied fields such as computational biology~\citep{schiebinger2019,bunne2021learning,bunne2022proximal,cao2022unified,bunne2023learning,klein2024entropicgromovwassersteinflow}, there is very limited work in the context of GeoAI, despite the roots of OT in transportation research. There are few exceptions; for example, \citet{roberts_gini-regularized_2017} experiment with spatio-temporal predictions in their methodological work on Gini-regularized OT. \citet{janati2020spatio} propose OT as a measure for the similarity of spatial time series with applications for clustering. \citet{liu2020geographic} coin the term ``geographical optimal transport'' for their application of OT for relocating geotagged tweets based on remote sensing data. More related work can be found in the realm of shared transport services, due to the obvious connection between OT and relocation. For example, \citet{treleaven_computing_2014} and \citet{treleaven_asymptotically_2013} employ the EMD with a road-map-based cost matrix for optimizing relocation in one-way car sharing, and \citet{qian2023exploring} measure the distance between bike sharing and public transport stations with the EMD. One form of unbalanced OT was proposed under the term ``graph-based equilibrium metric''~\citep{zhou2021graph} for measuring supply-demand discrepancies in ride sharing, with follow-up work that extends the metric by a more supply-sided view~\citep{chin2023unified}.

From the GIS perspective, related efforts primarily focus on incorporating spatial considerations into regression models, such as spatial error models~\citep{anselin2009spatial} (Spatial Lag in X, Spatial Autoregressive and Spatial Durbin Error Model), and spatially-explicit machine learning approaches~\citep{liu2022review}. Spatially-explicit models usually leverage spatial structure and autoregression, e.g. with Graph Neural Networks. Another stream of work aims to address spatial heterogeneity with local models such as Geographically Weighted Regression (GWR) and follow-ups~\citep{lesage2004family}, spatial Random Forests ~\citep{georganos2021geographical, sekulic_random_2020} or even Geographically Weighted Artificial Neural Networks~\citep{hagenauer2022geographically, du2020geographically}. However, these methods focus on integrating spatial knowledge into model \textit{training} and lack measures for evaluating model performance spatially. Spatial considerations in model \textit{evaluation} mainly involve measuring the residual spatial autocorrelation~\citep{gaspard2019residual} or to assess the generalization ability via spatial cross validation. We leverage OT to assess spatial aspects of the model performance. To the best of our knowledge, this is the first comprehensive analysis of OT for this purpose, demonstrated through synthetic data and real-world case studies. 


\section{Validation studies on synthetic data}\label{sec:synthetic}

To highlight the benefits of the proposed evaluation framework, we compare the OT error to the MSE as a standard error metric. We use synthetic data to demonstrate empirically that OT captures fundamentally different costs than the MSE, and relate these differences to spatial autocorrelation, a core concept in GIScience.

\subsection{Comparison to the mean squared error}

\begin{figure}[!htb]
    \centering
    \includegraphics[width=\textwidth]{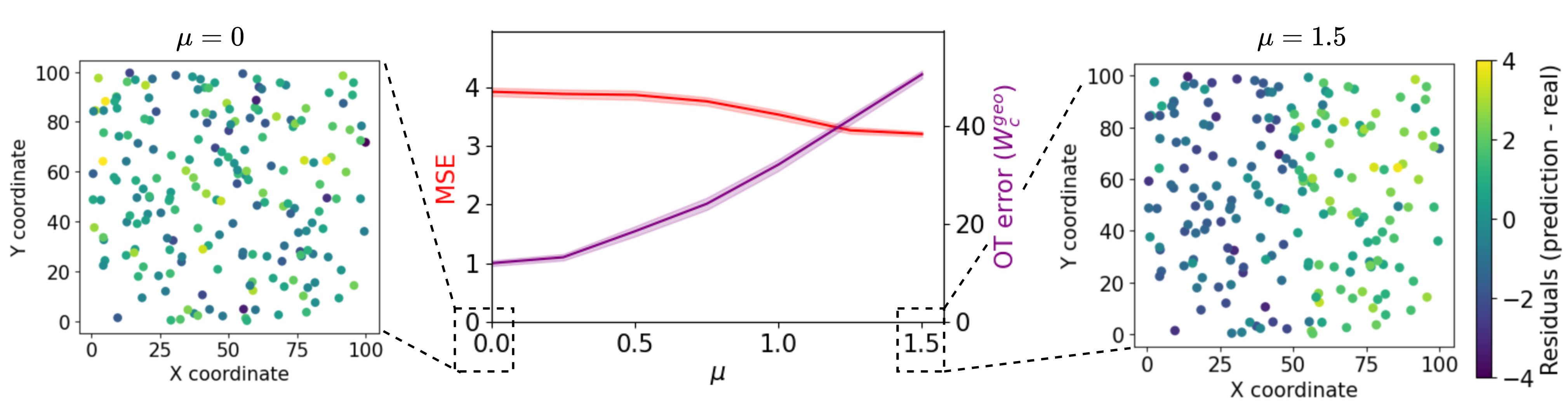}
    \caption{Comparison of MSE and OT error on synthetic data with increasingly unbalanced residuals ($\mu=0$: no imbalance, $\mu=1.5$: strong spatial imbalance). The imbalanced residuals lead to larger spatial costs, which is evident from the increasing OT error. There is also an evident relation of OT to spatial autocorrelation.}
    \label{fig:synthetic_comp}
\end{figure}

A synthetic scenario is designed to allow to systematically vary the spatial costs. Intuitively, the spatial costs, i.e. the transport costs to align ground truth and predictions, are higher if the residuals are unevenly distributed in space. To construct a simple scenario accordingly, we sample residuals from different distributions dependent on the x-coordinate of their location. Let $\*x_{i, 1}$ denote the first component of the location vector, i.e., its x-coordinate, and $\*x_{i, 2}$ the y-coordinate. The locations ($n=100$) are randomly sampled from a uniform distribution $\*x_{i,1}, \*x_{i,2}\sim U[0, 100] \ \ \forall i$, and the residuals from $\mathcal{N}(\mu, \sigma)$ for all locations with $\*x_{i, 1}<50$ and $\mathcal{N}(-\mu, \sigma)$ for $\*x_{i, 1}\geq 50$. 
The higher the $\mu$, the larger the \textit{spatial imbalance} of the residuals, i.e., the difference between the residuals at $\*x_{i,1}<50$ and the ones at $\*x_{i,1}\geq 50$. \autoref{fig:synthetic_comp} provides two examples, with $\mu=0$ corresponding to evenly distributed residuals, whereas $\mu=1.5$ results in an unbalanced spatial distribution. Such imbalance is very common in geospatial data due to spatial autocorrelation and spatial heterogeneity~\citep{zhang2009evaluation}. 

\autoref{fig:synthetic_comp} illustrates the OT error and the MSE for $\mu \in \{0, 0.5, 1, 1.5\}$. $\sigma$ was tuned to keep the average absolute value of the residuals constant. While the MSE remains constant, the OT error increases with the spatial imbalance, reflecting the increased transportation costs if the residuals are clustered in space, corresponding to large areas of oversupply distinct from areas of high demand. In turn, the transportation costs are lower if the errors are distributed randomly, since neighboring errors offset one another.

\FloatBarrier

\subsection{Relation to spatial autocorrelation of the residuals}

The synthetic experiment in \autoref{fig:synthetic_comp} shows that OT errors are higher if the residuals are clustered in space. This observation indicates a relation between the GeOT evaluation method and spatial autocorrelation. 
There are several measures to quantify spatial autocorrelation in a dataset, with global Moran's I~\citep{moran1950notes} arguably the most popular one. Moran's I is defined as:
\begin{gather}\label{eq:moransi}
    I = \frac{n}{\sum_{i,j=1}^n\*w_{ij}}\cdot\frac{\sum_{i,j=1}^n \*w_{ij} (\*v_i - \overline{\*v})(\*v_j - \overline{\*v})}{\sum_{i=1}^n (\*v_i - \overline{\*v})^2}
\end{gather}
where $\*v_i$ is the observation at the $i$-th location, $\overline{\*v}$ is the mean of all observed values, $n$ is the number of locations, and $\*w_{ij}$ is the (distance-based) weighting between two points. 

\begin{figure}[ht]
    \centering
    \begin{subfigure}[b]{0.38\textwidth}
    \includegraphics[width=\textwidth]{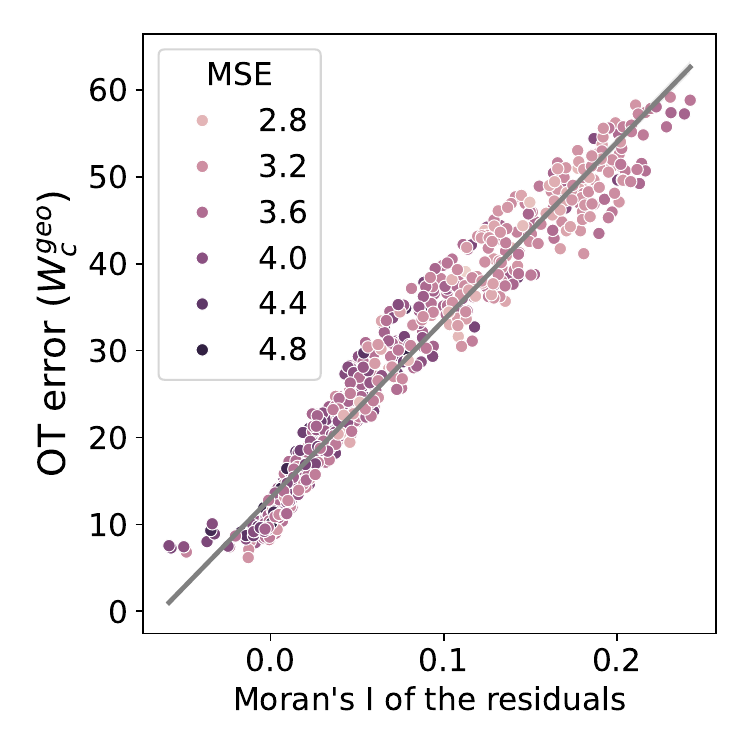}
    \caption{Empirical comparison}    
    \label{fig:ot_vs_moransi}
    \end{subfigure}
    \hfill
    \begin{subfigure}[b]{0.6\textwidth}
    \includegraphics[width=\textwidth]{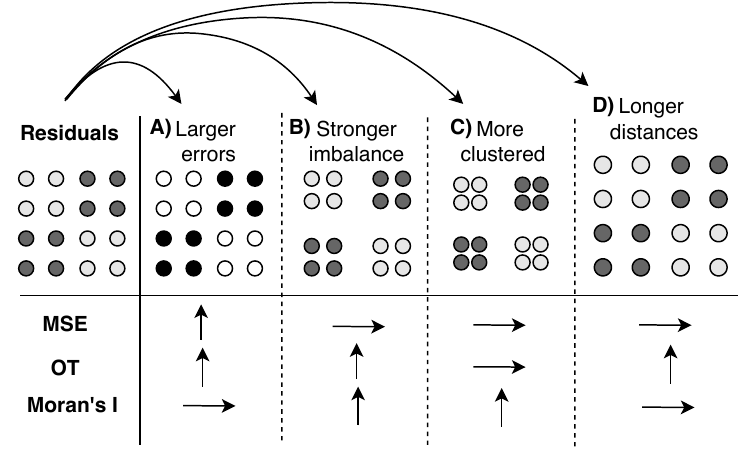}
    \caption{Differences between MSE, OT and Moran's I}
    \label{fig:moransi_schematic}
    \end{subfigure}
    \caption{Relation between the OT error and the spatial autocorrelation of the residuals, measured with Moran's I. Empirically, the OT error strongly correlates with Moran's I (\subref{fig:ot_vs_moransi}). Analytically we find that it combines standard error measures (A) with the ability to reflect spatial imbalance (B) and their distances (D). Since the OT error is computed as the minimal redistribution costs, it puts less focus on the similarity of neighboring points than Moran's I (C).}
\end{figure}

We empirically confirm the observed relation between the OT error and spatial autocorrelation by computing Moran's I on the synthetic data. For maximal comparability, we set the weights $\*w_{ij}$ to the negative costs, $\*w_{ij} = -\*C_{ij}$. Indeed, \autoref{fig:ot_vs_moransi} testifies a strong correlation of Moran's I and the OT error ($r=0.98$), largely independent from the MSE.

To understand this relation further, we examine the formula for Moran's I. 
We plug the residuals into \autoref{eq:moransi} by setting $\*v_i = \*o_i - \hat{\*o}_i$. To simplify $I$, we further assume \textit{balanced} OT, with $\sum_{i=1}^n \*o_i = \sum_{i=1}^n \hat{\*o}_i$. It follows that 
\begin{gather}
    \overline{\*v} = \frac{1}{n} \sum_{i=1}^n \*v_i = \frac{1}{n} \sum_{i=1}^n (\*o_i - \hat{\*o}_i) = \frac{1}{n} \sum_{i=1}^n \*o_i - \frac{1}{n} \sum_{i=1}^n \hat{\*o}_i = 0
\end{gather}
Thus, Moran's $I$ of the residuals becomes
\begin{gather}\label{eq:moransi_simple}
    I_{residuals} = \frac{n}{\sum_{i,j=1}^n \*w_{ij}} \frac{\sum_{i,j=1}^n \*w_{ij} (\*o_i - \hat{\*o}_i) (\*o_j - \hat{\*o}_j)}{\sum_i (\*o_i - \hat{\*o}_i)^2}
\end{gather}

While there is no direct theoretical relation between both measures (compare \autoref{eq:moransi_simple} to \autoref{eq:oterror}), the definition allows to understand their commonalities and differences, as illustrated in \autoref{fig:moransi_schematic}. First, due to the normalization factors in Moran's I, the absolute value of the residuals and the absolute distances do not matter (see \autoref{fig:moransi_schematic} A and D). The OT error increases with the error size and the distances. Furthermore, the spatial distribution of high and low residuals is reflected in both Moran's I and the OT error (\autoref{fig:moransi_schematic} B) - for the former as covariance ($(\*o_i - \hat{\*o}_i) (\*o_j - \hat{\*o}_j)$), and for the latter as transport mass; i.e., absolute differences. This explains the correlation observed in \autoref{fig:ot_vs_moransi}. However, OT computes the costs based on an explicit \textit{coupling} of residuals, while Moran's I measures the similarity between all points with positive $w_{ij}$. Thus, high similarities between \textit{nearby} points increase Moran's I (\autoref{fig:moransi_schematic}-C) whereas the OT error is driven  by the values of \textit{distant} points.

Thus, OT translates the rough indication of the spatial distribution of the residuals, provided by Moran's I, into a more precise measure of the associated operational costs. Although the GeOT framework cannot replace Moran's I or other measures for spatial autocorrelation, it provides a way to merge spatial considerations into standard error metrics for spatio-temporal data. 
More generally speaking, the close link between OT and such a fundamental concept as Moran's~I supports the value of OT in GIScience.

\FloatBarrier

\section{Case study: Evaluating bike sharing demand prediction with OT}\label{sec:case_study}

The GeOT framework is applicable to a wide range of applications, as shown in \autoref{tab:ot_applications}. The interpretation of the metric thereby depends on the application and on the definition of the cost matrix $\*C$. As a real-world example of a spatio-temporal forecasting problem, we utilize bike sharing demand prediction in the following. In this case, $W^{geo}_c$ can be interpreted as bike or user relocations that are necessary due to prediction errors. 
A public dataset is available from the BIXI bike sharing service in Montreal. 
The number of bike pickups at 458 stations is aggregated by hour and by station, following \citet{hulot2018towards}. A state-of-the-art time series prediction model, N-HiTS~\citep{challu_n-hits_2022}, is trained to predict the demand for the next five hours at any time point. The predictions are evaluated on a hundred time points from the test data period. For details on data preprocessing and model training, see \autoref{app:training} and \autoref{app:modeltraining} respectively.

\subsection{Relocation costs in bike sharing demand prediction}

\begin{figure}
    \centering
    \includegraphics[width=\linewidth]{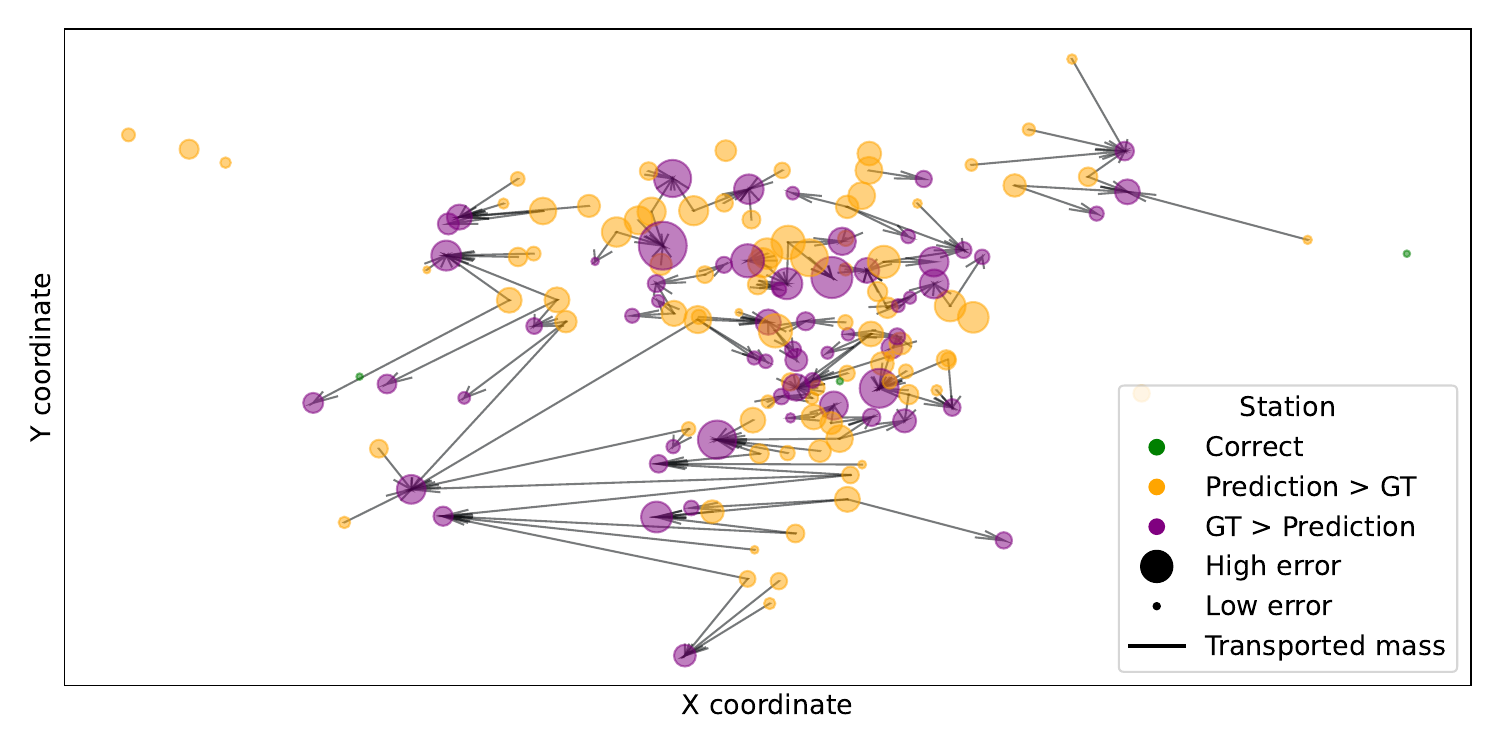}
    \caption{Transport map as computed with the GeOT framework. The goodness of the prediction is measured in terms of the relocation costs necessary to align the predictions with the real observations. Here, the difference between real and predicted bike sharing demand is shown, where mass is transported from bike sharing stations with overestimated demand (orange) to stations where the demand was underestimated (purple). In the example, the total spatial costs are rather low since most errors are balanced out with nearby points.}
    \label{fig:transport}
\end{figure}

First, we demonstrate the computation of the OT error using one example of bike-sharing demand predictions, for a single point in time. For visualization purposes, we subsample one third of the stations. \autoref{fig:transport} shows the spatial distribution of the residuals at these stations, highlighting, for example, a few stations with significantly underestimated demand (big purple circles) or an overestimation of bike-sharing demand in the bottom-left (orange points). 
Calculating the OT error involves computing $\*T^*$, the optimal transport matrix. We apply partial OT with $\phi=0$, essentially computing the difference between both distributions without penalizing the total difference of their masses (see \autoref{sec:partialot}). The arrows in \autoref{fig:transport} illustrate all nonzero cells of $\*T^*$, representing all required redistribution of mass to align the predictions with the true observations. The length of the arrows corresponds to the transport cost, since $\*C$ was set to the Euclidean distance between stations. In this example, most errors can be balanced out between neighboring stations, resulting in mass being relocated over short distances from prediction to ground truth. It is worth noting that a few errors are not balanced out since they are ignored through partial OT (see orange point in the bottom-left). The total spatial error corresponds to the sum of all arrow lengths when $\phi=0$, here $\mathrm{W}^{geo}_{\Tilde{c}, 0}=\sum_{i,j=1}^{n} \widetilde{\*C}_{ij} \*T^*_{ij}=58.91$. 

To interpret this error, assume that relocating one bicycle over one kilometer costs \$5. $W^{geo}_{\Tilde{c}, 0}$ represents the total relocation kilometers required to match the real bike-sharing demand with the predicted supply (apart from their total difference). Thus, the error of this prediction model would cost the bike-sharing service $58.91 \cdot \$5 = \$294.55$ if they needed to fully rebalance their supply to meet future demand. The GeOT framework's output could be integrated into more complex analysis tools specific to the company, such as considering the option of collecting and redistributing multiple bicycles simultaneously.

A major strength of OT is its flexibility to incorporate any arbitrary cost function, without requirements on the function's smoothness or other properties. This enables tailored application-specific analyses. In \autoref{app:user_relocation_time}, we showcase the OT error when setting the cost matrix to the user relocation time in bike sharing.

\subsection{Relation between OT errors, MSE and Moran's I}

To investigate the relation between OT error, MSE and Moran's I on real data, we compute these measures for all test samples of the bike sharing dataset. We expect the MSE to correlate with the OT error for real data, since a larger MSE typically goes along with greater mass redistribution. Indeed, we find 
a Pearson correlation of $r=0.38$ between the MSE and $\mathrm{W}^{geo}_c$ (with $\phi=0$). Similarly, there is a weak correlation between the OT error and Moran's I ($r=0.2$). The reasons for this weaker correlation, compared to the synthetic dataset (\autoref{fig:ot_vs_moransi}), are 1) the interdependence with the MSE, 2) the generally low spatial autocorrelation in the bike sharing data, 3) the low variance of the spatial autocorrelation in the data, which was controlled for the synthetic data. Together, MSE and Moran's I explain 58\% of the variance of the OT error when fitting a linear model, with coefficients of 1.04 and 0.94, respectively. This demonstrates that the GeOT framework combines both components into a unified metric while adding unique properties on top, by considering relocation distances explicitly.

\subsection{Comparability across scales}\label{sec:scales}

Research on spatio-temporal data oftentimes aggregates data across both space and time, leading to incomparable outcomes due to the Modifiable Areal Unit Problem (MAUP). The choice of aggregation size and method influences results, as observed in various analytical~\citep{gehlke1934certain, buzzelli2020modifiable} and predictive studies~\citep{smolak_impact_2021, smith_refined_2014}.
To address this issue in time series analysis, \citet{hyndman_optimal_2011} introduced hierarchical reconciliation~\citep{athanasopoulos2022evaluation, petropoulos_forecasting_2022} which improves consistency across space and time~\citep{kourentzes_cross-temporal_2019}. However, it still 
falls short in accounting for fuzzy groupings that arise from the ambiguity of clustering spatial locations. 


We argue that OT allows to compare results across scales and between different aggregation methods. Intuitively, aggregating the data in space decreases the error, since the clustered observations are less noisy. On the other hand, the \textit{utility} of the predictions is lower when they are not available on a fine-grained per-location level. In the following, we demonstrate how the GeOT framework can quantify this trade-off for the bike sharing data.
In bike sharing research, there is indeed a lack of comparability of previous work due to different aggregation schemes, ranging from single-station prediction~\citep{yang_mobility_2016, qiao_dynamic_2021} to various clustering schemes~\citep{hulot2018towards, shir_mobility_nodate, li_citywide_2020}. To capture this variety, we also aggregate the bike sharing data with several methods, namely 1) grouping by sociological or housing district\footnote{sociological districts from \url{https://www.donneesquebec.ca/recherche/dataset/vmtl-quartiers-sociologiques} and housing districts according to \url{https://www.donneesquebec.ca/recherche/dataset/vmtl-quartiers}}, 2) clustering with the KMeans algorithm (varying $k$), and 3) clustering with hierarchical (Agglomerative) clustering using different cutoffs. The bike sharing demand of a cluster is the sum of the demand of all its associated stations. One model is trained per configuration, where again the N-HiTS time series prediction model is used. The results are evaluated on the same test time points as before.

\begin{figure}[htb]
    \centering
    \includegraphics[width=\textwidth]{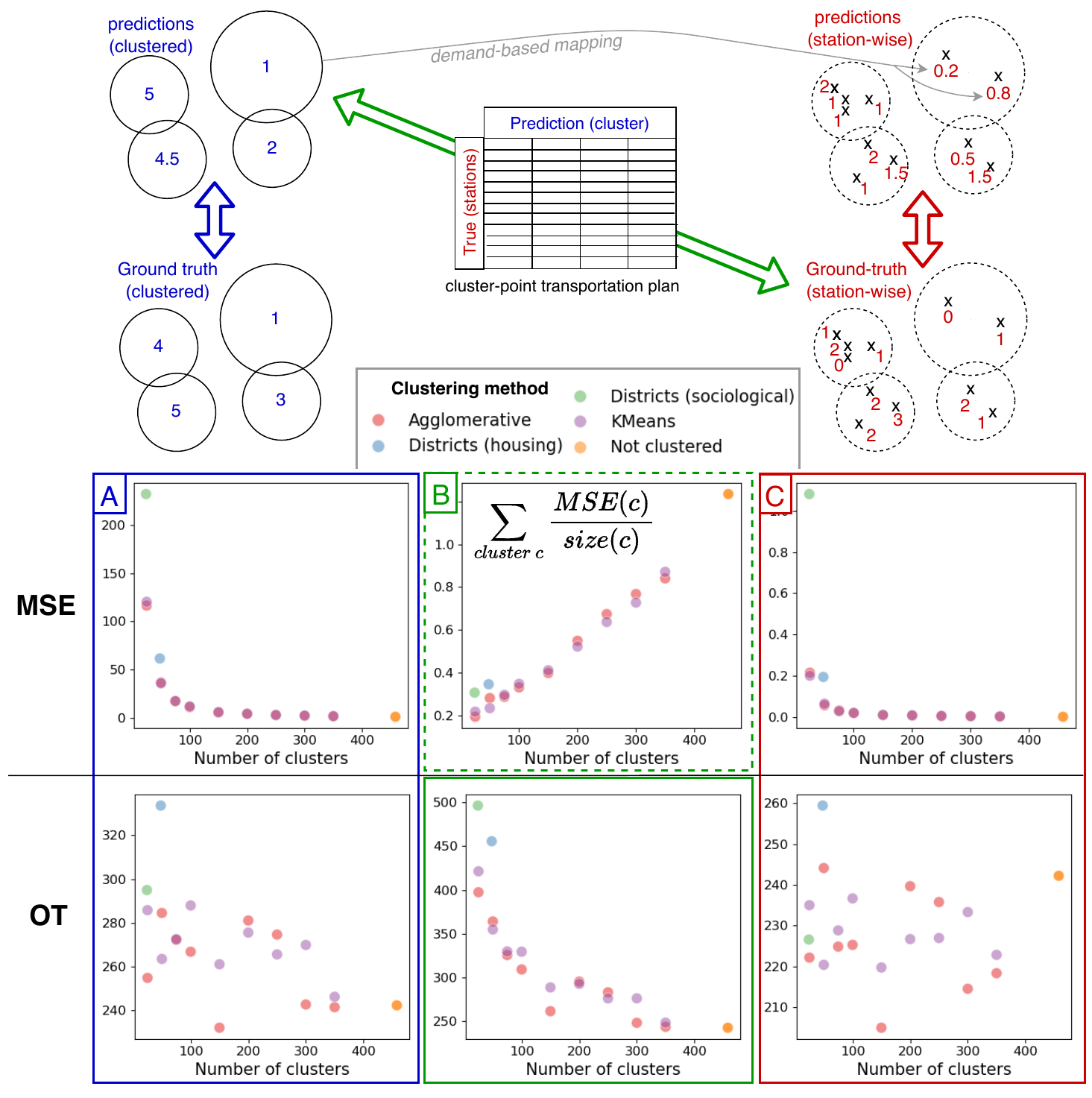}
    \caption{Comparing the prediction quality with MSE and OT error across spatial aggregation scales and clustering techniques. The MSE simply indicates higher errors for bigger clusters (A), or lower station-wise error with more data aggregation (B). Optimal transport allows for asymmetric cost matrices (green) to compute the costs for transporting from prediction-\textit{clusters} to the ground-truth-\textit{points} (B). In addition, OT takes into account the distances between clusters, providing a refined perspective on the optimal aggregation scale (A and C).}
    \label{fig:across_scales}
\end{figure}

As illustrated in \autoref{fig:across_scales}, we consider three evaluation methods: cluster-level comparison, evaluation of clustered predictions against point observations, and point-level errors. Cluster-level MSE, the standard approach taken in related work, decreases with more clusters (see \autoref{fig:across_scales}A) because each cluster contains fewer observations, typically resulting in lower errors. The OT error\footnote{Here, we use $\mathrm{W}^{geo}_{c, \phi}$ with $\phi$ equivalent to the 10\%-quantile of $\*C$ (see \autoref{app:choice_of_phi}), to model realistic business costs that arise mainly from redistribution but partly from a general over- or underestimation of bike sharing demand.} offers a different perspective as it accounts for distances between cluster centers, which increase when fewer clusters are used. Moreover, OT enables comparisons between cluster-predictions and station-level observations, as shown in \autoref{fig:across_scales} (green). In bike-sharing, for instance, cluster centers can be viewed as distribution hubs, and the OT error quantifies transport costs for redistributing bikes from hubs to stations. In this case, the OT error decreases with higher granularity (see \autoref{fig:across_scales}B), because it must redistribute the mass from the cluster centers to individual stations, which are further away from the hub if the cluster is larger. This insight enables balancing operational costs of additional hubs against reduced transport costs -- an analysis not possible with the MSE, which can only compare samples of the same size. One way to account for the clusters in the MSE is normalizing the prediction error by the cluster size (see dotted line in \autoref{fig:across_scales}B). In this case we can observe that larger clusters seem are easier to predict, probably because they exhibit more regular patterns. Finally, point-level errors (\autoref{fig:across_scales}C) are computed by allocating cluster predictions to stations based on their relative demand in the training set. The trends are similar but more pronounced than in \autoref{fig:across_scales}A. The MSE declines with the granularity, whereas the OT error balances accuracy with spatial granularity and achieves a minimum at 150 clusters found with Agglomerative clustering. In summary, the GeOT framework provides a refined evaluation across scales, capturing both absolute errors and their operational implications. When models are trained at multiple scales, the GeOT framework helps to select the optimal scale and aggregation method for each use case.

This concept can be extended to the \textit{temporal} scales and enable comparability between space and time. For instance, a prediction error at station A either make the bike sharing user relocate to another station, or wait at the station. In \autoref{app:space_and_time}, we provide an example with a spatio-temporal cost matrix, and show that the hierarchical N-HiTS model is particularly successful in terms of the resulting spatio-temporal OT error.

\FloatBarrier

\section{Case study: Evaluating spatial regression models with OT}\label{sec:interpolation}

As shown in \autoref{tab:ot_applications}, the GeOT framework is valuable not only for spatio-temporal forecasting but also for spatial interpolation and regression problems. To illustrate this, we examine a simple regression task: predicting Lung and Bronchus Cancer (LBC) mortality rates in new locations based on independent variables including PM25, SO2, and NO2 levels, poverty rates and smoking prevalance. A publicly available dataset for 2012\footnote{The data is available from https://zia207.github.io/geospatial-r-github.io/geographically-wighted-random-forest.html} includes data for 666 counties in the eastern United States. Mortality rates are expressed as the number of deaths per 100,000 people. Models are trained using 10-fold cross-validation, with all results reported on the test data. For OT cost calculations, we use the distances between county centroids.

\subsection{Model selection with OT}

To identify the best spatial regression model for this task with the OT error and other metrics, we compare linear regression (LR) to a Spatial Lag of X (SLX) model, a Spatial Autoregressive (SAR) model, Geographically Weighted Regression (GWR) and a Random Forest (RF) in \autoref{tab:model_comparison}. As expected, GWR and RF outperform simpler models in terms of RMSE. Analyzing Moran's I -- calculated using both standard 3-NN weights and weights identical to the negative costs $\*C_{ij}$ -- reveals that only GWR and RF effectively address spatial heterogeneity, whereas residuals from LR, SLX, and SAR predictions exhibit strong spatial correlation (see \autoref{tab:model_comparison}). The OT error combines the strengths of multiple evaluation measures: it highlights prediction errors (e.g., identifying SAR as inferior to other linear models) while accounting for the spatial distribution of residuals (e.g., showing that RF slightly outperforms GWR in minimizing residual autocorrelation). While model evaluation should consider multiple metrics, we emphasize GeOT as a valuable addition to standard evaluation frameworks.

\begin{table}[ht]
    \centering
    \resizebox{\textwidth}{!}{
    \begin{tabular}{lrrrrll}
    \toprule
    {} &   RMSE &  R-Squared &  Moran's I (3-NN) &  Moran's I (-C) & 
    $W^{geo}_{c, \phi}$ ($\phi$ low) & $W^{geo}_{c, \phi}$ ($\phi$ high)
    \\
    Method &        &            &                   &                 &                       &                        \\
    \midrule
    GWR    &  7.099 &      0.654 &             0.052 &           0.010 &               128k km &                320k km \\
    RF     &  7.278 &      0.636 &             0.036 &           0.006 &               124k km &                258k km \\
    LR     &  8.632 &      0.488 &             0.390 &           0.098 &               295k km &                315k km \\
    SLX    &  8.639 &      0.487 &             0.378 &           0.097 &               293k km &                309k km \\
    SAR    &  9.086 &      0.433 &             0.382 &          -2.039 &               433k km &               3510k km \\
    \bottomrule
    \end{tabular}
    }
    \caption{Model comparison with RMSE, Moran's I and OT error. GWR and RF perform best and can capture spatial heterogeneity. The OT error reflects the RMSE in its transport mass and the spatial distribution in its transport distances, thus showing a relation to Moran's I in this example.}
    \label{tab:model_comparison}
\end{table}

\subsection{Unpaired OT for evaluating spatial accuracy}\label{sec:unpairedot}

\begin{figure}[ht]
    \centering
    \begin{subfigure}[b]{0.32\textwidth}
        \includegraphics[width=\textwidth]{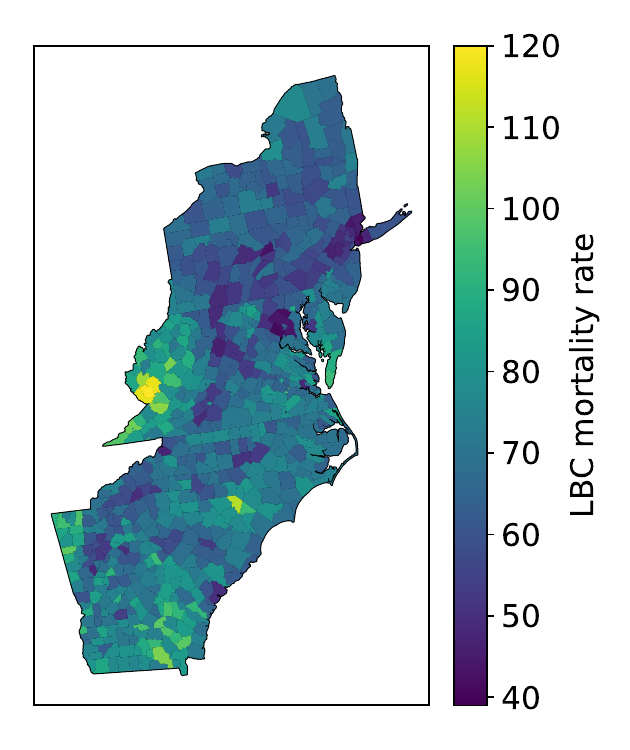}
        \caption{Observations}
        \label{fig:atlantic_gt}
    \end{subfigure}
    \begin{subfigure}[b]{0.32\textwidth}
        \includegraphics[width=\textwidth]{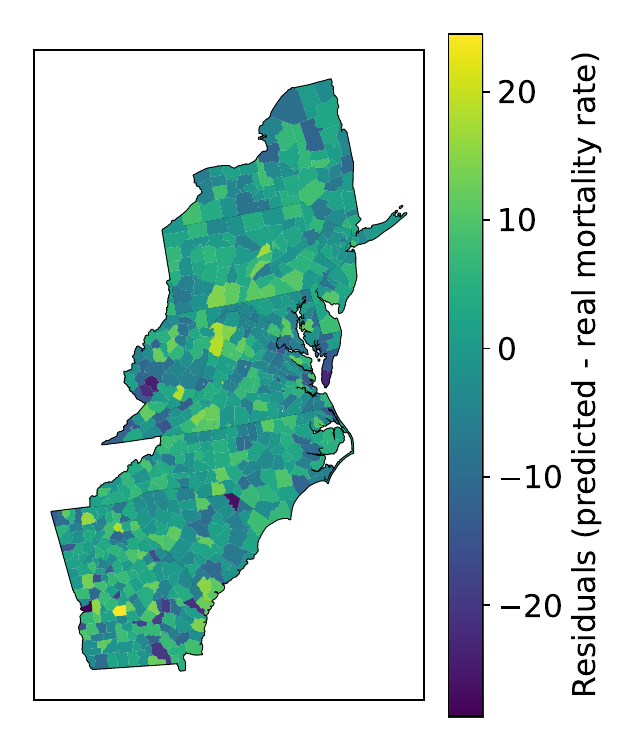}
        \caption{Residuals}
        \label{fig:atlantic_residuals}
    \end{subfigure}
    \begin{subfigure}[b]{0.32\textwidth}
        \includegraphics[width=\textwidth]{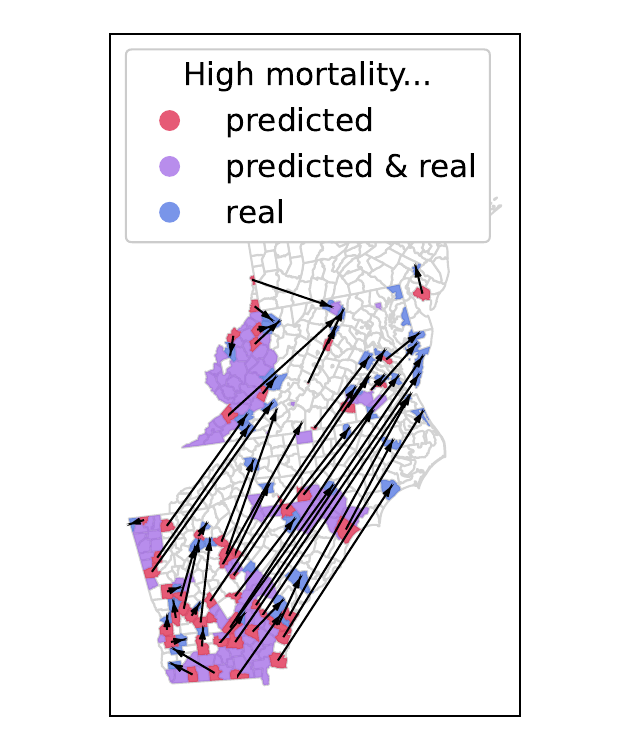}
        \caption{Optimal Transport plan}
        \label{fig:atlantic_transport}
    \end{subfigure}
    \caption{Evaluating the prediction of mortality rates (\subref{fig:atlantic_gt}) in unseen counties with GeOT. The residuals over all samples obtained with 10-fold cross validation are shown in (\subref{fig:atlantic_residuals}). The GeOT framework can be leveraged to compare the true and predicted locations of high LBC mortality. The transport plan (\subref{fig:atlantic_transport}) indicates a large spatial error in identifying the correct locations.}
    \label{fig:atlantic}
\end{figure}

In contrast to Moran's I, the GeOT framework provides an interpretable estimate of the spatial error, i.e., misplaced predictions. While so far we limited the analysis to the case where predicted and observed locations are the same, it is straightforward to use OT for computing the transport costs for two separate sets of points. To demonstrate this use case with the available dataset, we assume that a user wanted to identify counties with high mortality rates, defined as the ones where the rate lies in the highest quartile. \autoref{fig:atlantic_gt} shows the observed data, where bright colors correspond to high rates. \autoref{fig:atlantic_residuals} visualizes the residuals when predicting the rate with GWR. \autoref{fig:atlantic_transport} highlights only the counties of high predicted and / or high observed mortality rate. In this case, the distribution of the errors involves substantially large transport costs (black arrows). While high mortality rates occur also in the north-east, they were predicted predominantly in the south and west. This spatial error can lead to high opportunity costs of failed interventions, such as political measures that are executed in regions that are far away from the ones that are most affected.

\section{Experimental results for training models with an OT-based loss function}\label{sec:outlook}

Last, we investigate the use of an OT-based loss function to reduce spatial errors during training. This leverages the Sinkhorn loss as a differentiable relaxation of the Wasserstein distance, as introduced in \autoref{sec:sinkhorn} and \autoref{app:sinkhorn}.

\subsection{Data and experimental setup}

\begin{figure}
    \centering
    \begin{subfigure}[b]{0.32\textwidth}
        \includegraphics[width=\textwidth]{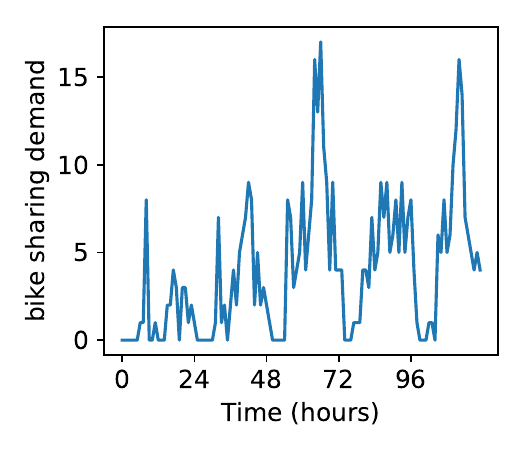}
        \caption{Bike sharing station}
        \label{fig:ts_bike}
    \end{subfigure}
    \begin{subfigure}[b]{0.32\textwidth}
        \includegraphics[width=\textwidth]{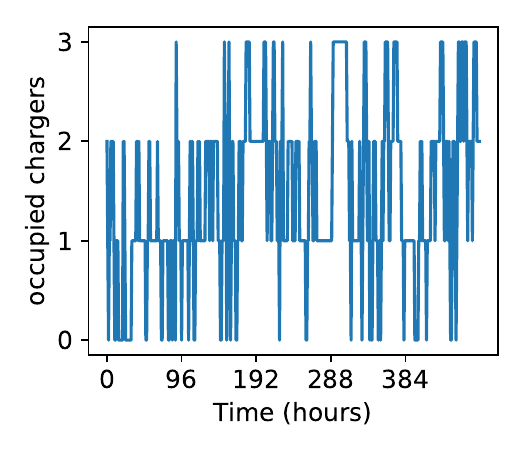}
        \caption{Charging station usage}
        \label{fig:ts_charging}
    \end{subfigure}
    \begin{subfigure}[b]{0.32\textwidth}
        \includegraphics[width=\textwidth]{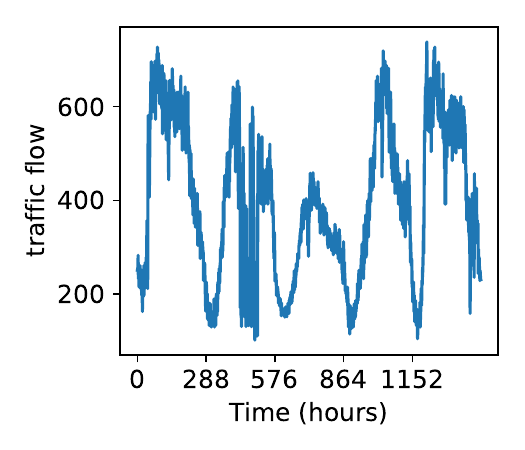}
        \caption{Traffic forecasting}
        \label{fig:ts_traffic}
    \end{subfigure}
    \caption{Example time series for the three tested applications. Each plot shows the time series at one location for an excerpt of five days. At the respective location, up to 17 bikes are picked up per hour (\subref{fig:ts_bike}), the traffic flow varies between 200 and 600 (\subref{fig:ts_traffic}), and usually 1-2 plugs of the charging station are occupied (\subref{fig:ts_charging}).}
    \label{fig:time_series}
\end{figure}

For a more comprehensive picture, we consider three applications: bike sharing demand, charging station occupancy, and traffic flow forecasting. A dataset on EV charging stations was published by \citet{amara2023forecasting}. It provides the charging station occupancy for 83 stations in France from July 2020 to February 2021 at a granularity of 15 minutes. For traffic forecasting, we use the popular PEMS dataset of traffic detectors on freeways in California. We take a small version of the dataset comprising 163 sensors and extract solely the traffic flow values that are available at a granularity of 5 min. \autoref{fig:time_series} shows a five-days excerpt of the respective time series for the three applications. We train the N-HiTS model with the Sinkhorn divergence as the loss function and compare the results to the ones obtained previously when training with an MSE loss. 
All models were trained to predict the demand for the next five time steps. For details on the data sources, preprocessing and the model training, see \autoref{app:training} and \autoref{app:modeltraining}. 
While OT has been used for evaluating spatio-temporal forecasts~\citep{roberts_gini-regularized_2017}, to the best of our knowledge, this is the first attempt to improve forecasts of geospatial data with a Sinkhorn divergence function.

\subsection{Comparison of models trained with Sinkhorn loss and MSE loss}

The trained models are evaluated on test data in terms of the MSE, the balanced OT error (scaling $Y$ and $\hat{Y}$ to have equal sums), and $\mathrm{W}^{geo}_c$ with small and large $\phi$ (see \autoref{app:choice_of_phi} for further analysis on the choice of $\phi$ and its effect on $\mathrm{W}^{geo}_c$). The cost matrix $\*C$ was set to the Euclidean distance between stations in km. This results in an interpretable model performance. For example, a relocation effort of around 135.7km is required in total to align the predicted bike sharing demand with the true one.  \autoref{tab:ot_loss_results} demonstrates that $W^{geo}_{c, \phi}$ can be reduced to some extent when training with the Sinkhorn loss. For the bike sharing data, all OT-based metrics are substantially improved, with minimal impact on the MSE. For the charging station application, $\mathrm{W}^{geo}_c$ decreases, but the OT error with higher $\phi$ is not improved. This may be due to a better reduction of the total error $\delta$ when using the MSE loss. For traffic forecasting, all OT-based metrics improve, but the MSE increases more compared to the other applications.

\begin{table}[htb]
\centering
\resizebox{\textwidth}{!}{
\begin{tabular}{llrrcc}
\toprule
 &  & MSE & $\mathrm{W}^{geo}_c$ & $W^{geo}_{c, \phi}$ ($\phi$ low) & $W^{geo}_{c, \phi}$ ($\phi$ high) \\
Application & Loss function & & &  &  \\
\toprule
Bike sharing demand & OT (Sinkhorn) loss & 1.26 & 135.7 & 195.7 & 1733.8 \\
 & MSE loss & 1.24 & 161.5 & 242.2 & 2406.1 \\
\midrule
Charging station & OT (Sinkhorn) loss & 0.35 & 30.7 & 30.8 & 87.0 \\
occupancy & MSE loss & 0.34 & 32.7 & 30.7 & 81.1 \\
\midrule
Traffic flow & OT (Sinkhorn) loss & 876.63 & 1629.3 & 1565.9 & 5558.4 \\
 & MSE loss & 852.53 & 1639.3 & 1598.2 & 5892.3 \\
\bottomrule
\end{tabular}
}
\caption{Results when training with an OT-based loss function. At minor increase of the MSE, OT-based metrics can be decreased substantially; e.g., from 161.5 to 135.7km bike relocation cost.}
\label{tab:ot_loss_results}
\end{table}

\begin{figure}[bht]
    \centering
    \includegraphics[width=\textwidth]{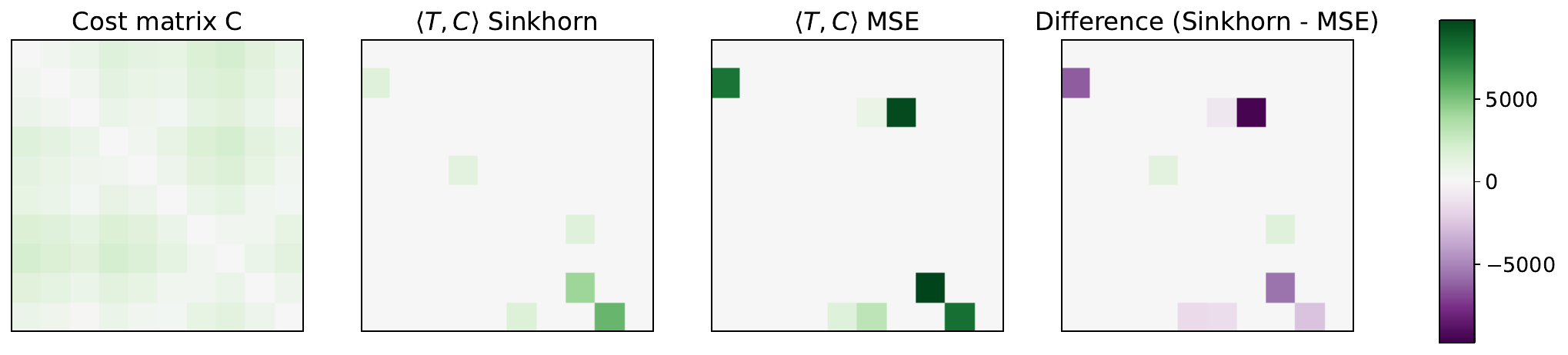}
    \caption{Transportation costs for aligning traffic flow predictions with the ground truths. The plot compares the results between a model trained with an MSE loss to a model trained with the Sinkhorn divergence. $\odot$ denotes the element-wise multiplication of two matrices; here the OT matrix $T$ and the cost matrix $\*C$.}
    \label{fig:transition}
\end{figure}

\autoref{fig:transition} illustrates transportation matrices for one time point of the traffic prediction task. For visualization purposes, only a subset of the traffic sensors is shown, selected via spectral clustering on the distance matrix and randomly choosing one cluster. Training with the Sinkhorn loss results in lower transportation costs in this example. 

In sum, these experiments demonstrate that the OT-based metrics presented in this paper can be minimized through tailored loss functions. However, there is a trade-off between enhancing the spatial distribution of predictions and maintaining standard MSE performance across applications.  Moreover, the higher computational complexity of the Sinkhorn loss should be considered; analyzed in detail in \autoref{app:runtime}. Training with the Sinkhorn loss can nevertheless be beneficial in scenarios where 1) the spatial distribution of the error plays a significant role and the OT error has a clear real-world interpretation, or 2) when methods suffer from residual autocorrelation.


\section{Discussion}\label{sec:discussion}

Optimal Transport holds significant potential to incorporate spatial considerations into both the evaluation and training of GeoAI models. First, the GeOT framework provides \textit{interpretable} error metrics linked to operational costs, making it valuable for practical applications. Second, GeOT inherently accounts for \textit{spatial} relationships, serving as a spatially-explicit metric that evaluates prediction errors in the context of their locality. Last, its \textit{flexibility} allows for customization to specific applications through the use of tailored cost functions. 

The decision how to implement OT-based metrics should consider the specific prediction problem and the use of predictions in downstream tasks. Its relevance varies across applications; for instance, it is particularly useful for evaluating predictions of transport demand, but less so in traffic forecasting, where the spatial distribution of errors is less critical. To determine its applicability, we recommend to construct a cost matrix with a certain real-world meaning (e.g., relocations, mission planning, signal control, interventions). If there exists such cost matrix with obvious real-world relevance, an evaluation with the GeOT framework is suitable. At the same time, it is important not to overinterpret the results in light of the complexity of real-world applications. For instance, in the bike sharing example, OT-based relocation costs abstractly represent the system operator's costs but do not account for complexities like batched relocations, employee schedules, or other operational factors. Thus, OT errors are rather an abstract approximation of application-specific costs that arise based on the spatial distribution of the errors, oftentimes due to spatial heterogeneity. In general, the GeOT framework should be seen as an important new component of model evaluation rather than a sole new metric that replaces the MSE or other standard metrics.

While GeOT is generally applicable to spatio-temporal prediction problems, there are further limitations and challenges. First, OT cannot deal with negative values; however, this problem can usually be resolved with a simple scaling of all values. 
Secondly, a pivotal choice in our framework concerns the value of $\phi$, i.e. the costs for importing or exporting mass in partial OT. A smaller $\phi$ emphasizes accuracy in spatial distribution over the minimization of total error. Selectively adjusting $\phi$ by application poses a risk for cherry-picking; i.e. setting $\phi$ to the value with the best results. While further analysis is required to gain a better understanding of the dependency of the error on $\phi$, our preliminary tests in \autoref{app:choice_of_phi} suggest to set $\phi$ to the maximum value of the cost matrix. Finally, calculating the OT error is computationally more expensive than other metrics. While modern LP solvers and the Sinkhorn algorithm generally offset these concerns, challenges may arise when handling datasets with numerous locations, such as remote sensing imagery. Finally, experiments with the Sinkhorn loss showed that decreasing the OT error can come at the cost of higher MSE. Our work provides guidelines on possible research avenues and outlines general advantages over non-spatial metrics, but further research is required to substantiate these advantages in specific real-world applications.

\section{Conclusion}\label{sec:conclusion}

In compliance with Tobler's law that near things are more related than distant things~\citep{tobler1970computer}, this paper introduces the GeOT framework, a spatially-explicit method for evaluating spatio-temporal predictions. It leverages Optimal Transport theory, an established mathematical field with a strong theoretical foundation.
Our experiments on synthetic and real data demonstrated the value of OT for geospatial contexts, showcasing its real-world meaning due to its quantification of the potential cost reductions, its ability to incorporate application-specific criteria with arbitrary cost matrices, its relation to key geographical concepts such as spatial autocorrelation, and its utility as a loss function. 

There are numerous paths for future work. A first step to establish GeOT as an evaluation framework in specific GIScience fields involves reviewing existing methodologies and comparing them with respect to the OT error. Additionally, we have showcased and briefly demonstrated several promising directions, such as integrating space and time into a unified OT metric, which should be investigated in more detail. 
Moreover, the combination of OT and GIScience is not a one-way street, since some of the presented ideas are interesting for a more general machine learning audience. For example, the proposed concept of space-time-cost matrices implies that there is a use case in general time series analysis beyond spatial data. In any multi-step time series prediction task, OT could be used to quantify the temporal prediction inaccuracies. Last, our experiments with the Sinkhorn divergence show the potential of a wider application of OT-based loss functions with location-based cost matrices, such as in remote sensing. 

Ultimately, we not only aim to convince GIScientists and GeoAI researchers of the value of OT in the field, but stimulate further research into its application-specific uses, domain-independent applications, and into enhancing model training with OT-based approaches.

\section*{Acknowledgements}
We would like to thank Thomas Spanninger for the fruitful discussions and valuable feedback. Furthermore, many thanks to Thomas Klein and Jannis Born for reviewing a prior version of this paper. 
This project is part of the E-Bike City project funded by the Department of Civil and Environmental Engineering (D-BAUG) at ETH Zurich and the Swiss Federal Office of Energy (BFE).

\section*{Declaration of interest} The authors report there are no competing interests to declare.

\section*{Author contributions} 
NW: Conceptualization, Methodology, Software, Validation, Formal Analysis, Investigation, Data Curation, Writing -- Original Draft, Writing -- Review and Editing, Visualization\\
TU: Formal Analysis, Writing -- Original Draft, Writing -- Review and Editing \\
MR: Conceptualization, Writing -- Review and Editing, Supervision, Funding Acquisition

\section*{Data and Codes Availability Statement} All code is publicly available at \url{https://github.com/mie-lab/geospatial_optimal_transport/}, 
and, as a tutorial version, at \url{https://github.com/mie-lab/geospatialOT}.
All datasets are publicly available as detailed in \autoref{app:training}.

\section*{Generative Artificial Intelligence}

ChatGPT version 4o was used for language improvements.





\bibliographystyle{tfv}
\bibliography{references}

\newpage
\appendix
\begin{center}
  {\LARGE \bfseries Appendix}
\end{center}
\section{Applications}\label{app:applications}

The proposed framework is applicable to any spatial machine learning problem where values are predicted for a set of locations. OT should be included in the model evaluation whenever the spatial \textit{displacement} of the predictions plays a role; i.e., when the spatial distribution of the errors has real-world implications. The exact application of OT varies depending on the prediction problem. In this section, we expand on the applications listed in \autoref{tab:ot_applications}. Specifically, we outline the applicability of OT by distinguishing three kinds of applications: vision-based GeoAI (essentially segmentation of raster data), point-data time series analysis, and spatial interpolation problems.

\subsection{Vision-based GeoAI}

Machine learning is used in many geospatial applications, but has arguably brought the most significant advances to problems involving \textit{raster} data. The raster format allows to leverage Convolutional Neural Networks (CNNs) that were developed for image processing but also show impressive performance in classifying, segmenting or regressing geographical raster data. Examples include the analysis of remote sensing data, e.g., for land use classification~\citep{alem2020deep} or for glacier retreat prediction, but also weather forecasting~\citep{kareem2021evaluation} or for predicting sociodemographics in rasterized population data.
In remote sensing applications, the spatial distribution of the prediction errors oftentimes plays an important role, for instance in the locality of weather phenomena or the extend of wildfire spread~\citep{radke2019firecast,salis2016predicting}. 
Relocating resources, e.g. for extinguishing fire, costs valuable time. OT can quantify such negative effects of prediction errors. 

For raster data, the locations $l_i$ correspond to the raster cells or pixels, and the cost matrix $\*C$ to the pairwise pixel distances, or the operational costs to relocate resources from one pixel to another. 
For instance, consider the problem of predicting wildfire spread by classifying pixels as ``fire'' or ``no fire''. With the proposed framework, the locations of predicted ``fire'' pixels can be compared to the true fire extent by means of optimal transport plans. The OT error is higher if the wildfire spreads into an entirely different direction than expected. In contrast, standard accuracy metrics only depend on the pixel counts and do not reflect spatial properties.

\subsection{Point-data time series}

The focus of this paper is on spatio-temporal prediction problems where the aim is to predict the future observations at given locations. Examples are abundant in the transportation field, for instance including research on bike sharing demand~\citep{shir_mobility_nodate, hulot2018towards, yang_mobility_2016, qiao_dynamic_2021, liu2015station, li_citywide_2020}, one-way car sharing flow~\citep{zhu_multistep_2019}, e-scooter fleet utilization~\citep{abouelela_exploring_2023} or generally shared services~\citep{xiao2023graph}. In these fields, ML models are applied to predict transport demand at multiple stations, with the goal of improving efficiency. 

The OT error allows to quantify the model's ability to predict the spatial demand distribution in an interpretable manner.
For instance, assume the demand at location $a$ is underestimated while it is overestimated at $b$. In the OT framework, this leads to a positive value in the OT plan, e.g. $T_{b,a}>0$. The transport of mass in $T$ can be interpreted as users having to walk from station $a$ to station $b$ to pick up a bike, since the demand at $a$ was underestimated and an insufficient number of bikes is available there. In another interpretation, the flow can be seen as the number of bikes that need to be relocated from $b$ to $a$ at the end of the day to balance out the bike availability. It is worth noting that the business costs of operations in a shared system are far more complex; however, the OT metric adds a spatial aspect to the model evaluation, yielding a better estimate of business costs than non-spatial (standard) error metrics.

However, OT is not restricted to applications that entail relocation operations. For example, consider energy demand prediction, an important research direction in light of the challenging integration of renewable energies. 
Power system operations and the adoption of electric vehicles (EV) can be supported by predicting the occupancy rate of EV charging stations. The occupancy is the number of plugs at the station that are currently in use. Predictions are relevant for navigating vehicles to the closest stations with available plugs. In turn, prediction errors translate to relocation costs for EV drivers who find a charger unexpectedly occupied. The OT cost thus reflects that real-world costs for drivers are higher the further away the next available station is. 

Last, another example is traffic forecasting, a popular branch in GeoAI. Traffic forecasts are relevant for operating signals and for navigation purposes. Prediction errors of traffic forecasting models are more severe if the spatial distribution of traffic is not estimated correctly; i.e., if traffic is expected far from the location where it actually occurs. Thus, OT is useful as a spatial metric in traffic forecasting even though it is not directly interpretable in terms of operational costs. In particular, OT-based evaluation can yield insights into the actual advantages of novel model architectures that are being proposed as spatially-explicit neural networks.

\subsection{Spatial interpolation and regression}

OT can also be applied to regression problems without a temporal component, for example, inferring housing prices from features such as the property size and its proximity to the city center, or predicting deforestation risks in the rain forest from covariates such as sociodemographics, spatial features, and economic information. In this case, the framework introduced in \autoref{sec:methods} can be applied in the same manner as for spatio-temporal time series, since the OT error is computed per time step in any case.

\section{Data and preprocessing}\label{app:training}

The bike sharing dataset was downloaded from Kaggle\footnote{\url{https://www.kaggle.com/datasets/aubertsigouin/biximtl}} and restricted to the period from 15th of April to 15th of November 2014, since the service is closed in winter, leading to large gaps in the time series across years. Only stations with missing coordinates or maintenance stations were removed. 

The charging station occupancy dataset was published by \citet{amara2023forecasting} in the context of the ``Smarter Mobility Data Challenge''. Each charging station has three plugs and the challenge is to classify the state of each plug as ``available'', ``charging'', ``passive'' (plug is connected to a car that is fully charged) and ``other'' (out of order). Here, we frame the task as a regression problem of predicting the fraction of plugs that are occupied, i.e., \textit{charging} or \textit{passive}. The forecasts could help to estimate the energy demand and to facilitate planning of charging stops for owners of electric vehicles.

The data is given at a granularity of 15 minutes from 3rd of July 2020 to 18th of February 2021. The time series is comparably sparse, since a station on average has no plugs in use 61\% of the time; one out of three plugs in use by 27\%, and only 2.1\% where all three plugs are used. From 2020-10-22 onwards, there is also a considerable number of missing data, amounting to 8\% of missing information on the number of chargers in use. We execute the preprocessing pipeline\footnote{Available on GitHub:\url{https://github.com/arthur-75/Smarter-Mobility-Data-Challenge}} of the winning team of the ``Smarter Mobility Data Challenge'', who employ exponential moving weighted average to fill missing values. We further removed stations with no charging activity, leaving 83 charging stations. Finally, we scale all values by dividing by 3 for training the model.

The PEMS traffic dataset is published by the California Department of Transportation\footnote{\url{https://pems.dot.ca.gov}} and was downloaded from GitHub\footnote{\url{https://github.com/guoshnBJTU/ASTGNN/tree/main/data/PEMS08}} where it was published by \citet{guo2021learning}. It provides traffic flow, traffic occupancy rate, and traffic speed at each sensors in five minute intervals for 62 days (July to August in 2016). We predict only traffic flow. Furthermore, there data includes the spatial distances between certain pairs of sensors. We construct the cost matrix by computing the all-pairs shortest paths in an undirected graph that was built from the given distances. 

\section{Model training}\label{app:modeltraining}

In all cases, we train an established time series prediction model, N-HiTS~\citep{challu_n-hits_2022}, implemented in the \texttt{darts} library~\citep{herzen2022darts}. The model was chosen since it outperformed other common approaches such as Exponential Smoothing, LightGBM~\citep{ke2017lightgbm} or XGBoost in our initial experiments. 

The model is trained for 100 epochs with early stopping. The learning rate was set to $1e^{-5}$. The time series was treated as multivariate data with one variable per bike sharing station or charging station. A lag of 24 is used to learn daily patterns, and the hour and weekday are provided as past covariates. The number of stacks in the N-HiTS model was set to 3. The number of output time steps corresponds to our forecast horizon of five time steps. For evaluation, we draw 100 samples from the test data (last 10\% of the time series) and predict the next five time steps based on the respectively preceding time series, without re-training the model. For further implementation details, we refer to our source code.

\section{Partial OT as a combination of distributional and total costs}\label{app:choice_of_phi}

\begin{figure}[t!b]
    \centering
    \includegraphics[width=0.7\textwidth]{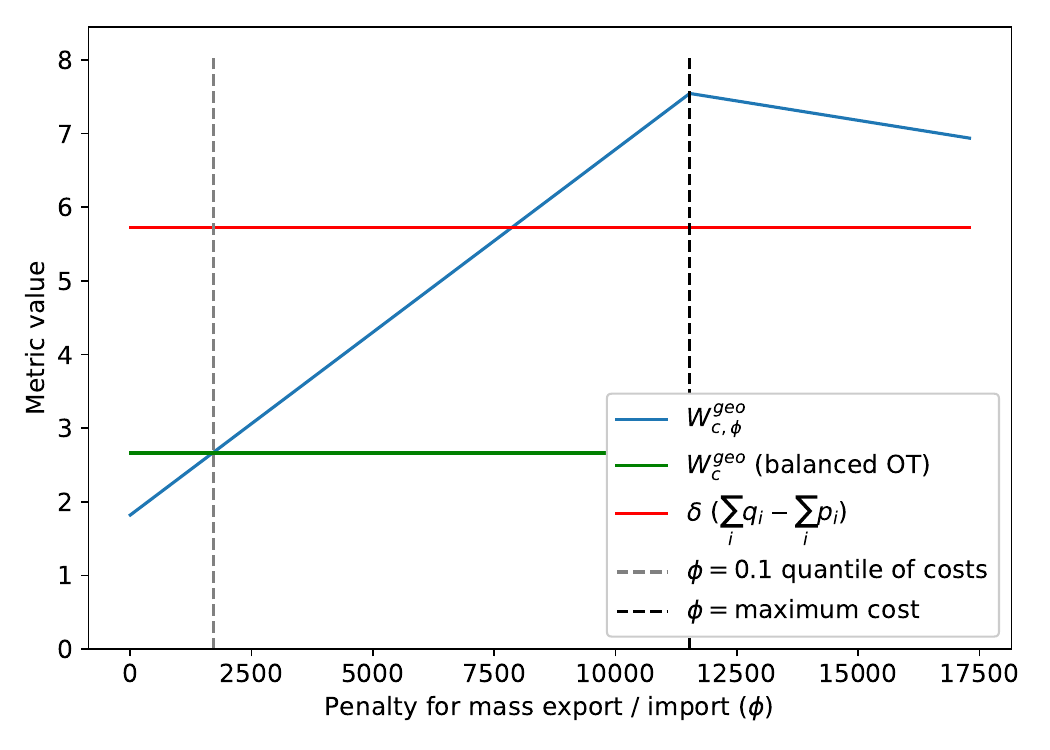}
    \caption{Relation of $W^{geo}_{\Tilde{c}, \phi}$ to $\phi$ in one synthetic example}
    \label{fig:uot_example}
\end{figure}

Intuitively, partial OT strikes a balance between balanced OT (measuring the mismatch between the predicted and true distribution) and the total error $\delta$ (mismatch between the sum of predicted and the sum of observed values). The weighting between both depends on $\phi$. In \autoref{tab:ot_loss_results}, we reported the results for the partial OT error with low $\phi$, specifically setting $\phi$ to the 0.1-quantile of all pairwise costs $\*C_{ij}$, and high $\phi$, where $\phi$ is set to the maximum of the cost matrix $\phi = \max_{ij} \*C_{ij}$. The reasoning of these parameter settings is illustrated in \autoref{fig:uot_example}, showing the OT error of the model trained on predicting charging data occupancy with a Sinkhorn loss. In particular, \autoref{fig:uot_example} illustrates the dependence of $W^{geo}_{\Tilde{c}, \phi}$ on $\phi$. For ensuring comparability of $W^{geo}_{c}$ and $W^{geo}_{\Tilde{c}, \phi}$, the extended cost matrix $\widetilde{\*C}$ was normalized by its maximum for this illustration. 

We observe that the $W^{geo}_{\Tilde{c}, \phi}$ approximately corresponds to $W^{geo}_{c}$ when $\phi$ is set to the 0.1-quantile of $\*C$ (intersection of green and blue lines in \autoref{fig:uot_example}). This observation is consistent for synthetic data as well as the bike sharing dataset. The reason is that with $\phi=0$, only the spatial distribution would be penalized, but some mass could be imported to / exported from arbitrary locations for free. Thus, $\phi=0$ leads to lower errors than $W^{geo}_{c}$.

On the other hand, for $\phi\longrightarrow\infty$, all entries of $\*C$ become zero except for the last row and column which is 1, since all values are divided by $\phi$ when normalizing by the maximum.

Thus, $W^{geo}_{\Tilde{c}, \phi}$ converges to $\delta$ for large $\phi$ (blue line approaching red line). When $\phi=\max_{ij} \*C_{ij}$, the partial OT error is maximal since the balanced error, $W^{geo}_{c}$, is combined with $\delta$.

\FloatBarrier

\section{Flexibility to define application-specific cost matrices}\label{app:user_relocation_time}

\begin{figure}
    \centering
    \begin{subfigure}[b]{0.36\textwidth}
        \includegraphics[width=\textwidth]{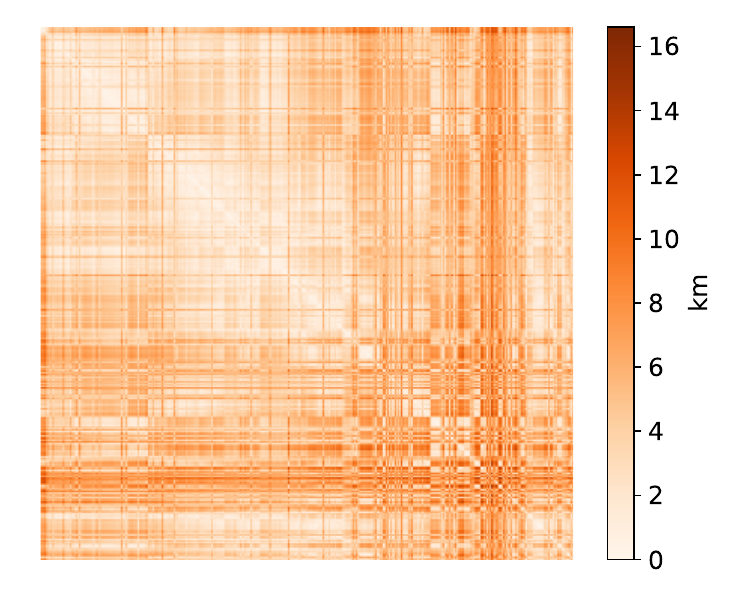}
        \caption{Euclidean cost matrix}
        \label{fig:cost_eucl}
    \end{subfigure}
    \begin{subfigure}[b]{0.29\textwidth}
        \includegraphics[width=\textwidth]{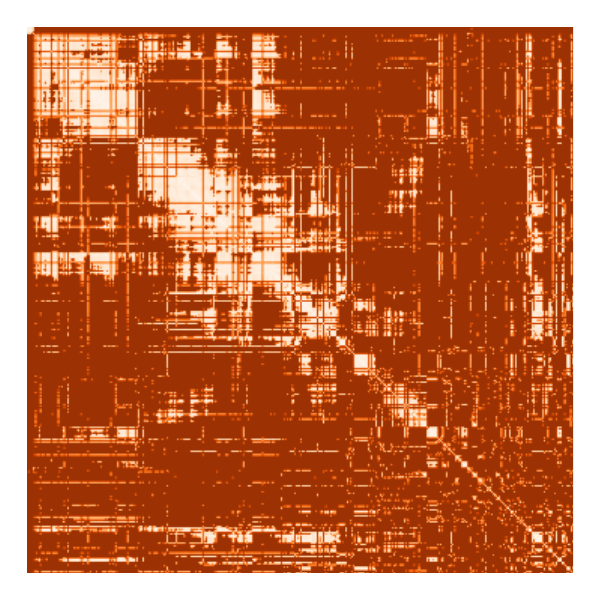}
        \caption{Map-matched costs}
        \label{fig:cost_mapmatched}
    \end{subfigure}
    \begin{subfigure}[b]{0.32\textwidth}
        \includegraphics[width=\textwidth]{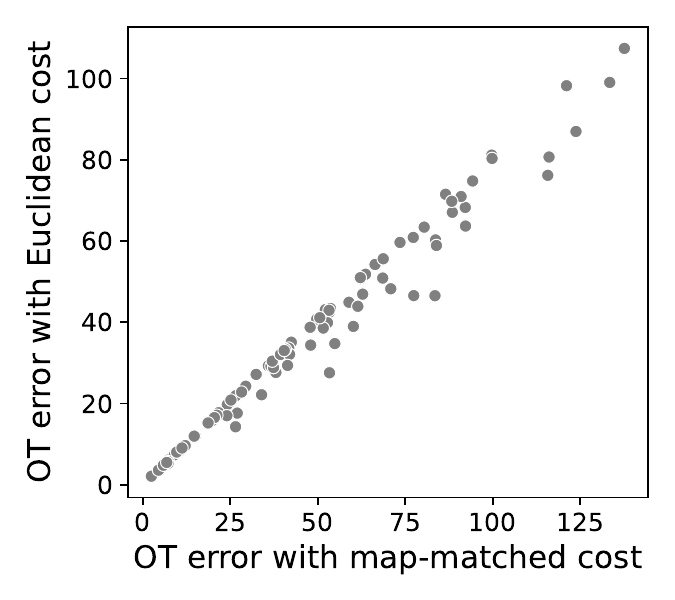}
        \caption{Relation of OT error}
        \label{fig:relation_mapmatched_eucl}
    \end{subfigure}
    \caption{Dependence of the OT error on the given cost matrix. The GeOT framework can incorporate application-dependent cost matrices. For example, a map-matched cost matrix (with cutoff) leads to larger errors than a simple Euclidean-distance-based cost matrix.}
    \label{fig:mapmatched}
\end{figure}

In the bike sharing example, we could also interpret prediction errors as \textit{users} having to walk from one station to another. In this case, it is more sensible to set $\*C$ to map-matched walking distances. On top of that, the effort of users is not linear, since users would dismiss bike sharing as a transport option when no bike is available in any feasible distance. We construct a cost matrix accordingly, where the cost between pairs of locations that are closer than 2km corresponds to the map-matched walking distance, and the cost between all other pairs is set to 15km to express the high costs of loosing customers when no bike is available anywhere nearby. 
\autoref{fig:relation_mapmatched_eucl} provides the new cost matrix in comparison to a simple Euclidean-distance based matrix, and shows the resulting OT errors $W^{geo}_{\Tilde{c}, 0}$ for the test set, i.e., the predictions from the N-HiTS model for 100 randomly selected time points. The errors correlate, but the map-matched cost matrix generally results in larger relocation costs and the error diverges significantly from the previous result for certain samples.

\section{Comparability between space and time}\label{app:space_and_time}

Time and space are currently treated as independent dimensions when evaluating spatio-temporal prediction models. For example, when predicting bike sharing demand for the next five hours, the errors are usually reported as averages over time and over locations. To the best of our knowledge, there are no methods to compare inaccuracies in time (e.g., predicting high demand for a later time than when it actually emerges) to inaccuracies in space (e.g., overestimating the demand at station A while underestimating station B). We suggest to enable the joint evaluation of spatial and temporal errors with OT, by translating relocation costs into relocation \textit{time}. The proposed costs are visually explained in Figure~\ref{fig:spacetime} and detailed in the following. We first define the cost matrix $\*C$ in terms of time instead of distance. For the bike sharing example, we derive the walking time from the spatial distances of locations by assuming a walking speed of 5km/h. 
We now extend the spatial cost matrix to a space-time cost matrix $\Psi$. Let $\Psi_{(i,t_k),(j,t_l)}$ be the time to relocate from the $i$-th location at time $t_k$ to the $j$-th location at time $t_l$. Dependent to the specific application, it must be decided how to penalize \textit{temporal} errors - what is the cost for $\Psi_{(i,t_k),(i,t_l)}$, i.e. the transport of mass from one time point to another? A simple option is to set the temporal relocation cost to $|t_k - t_l|$ (see Figure~\ref{fig:spacetime}). This could be viewed as a \textit{waiting} time until a bike becomes available. Finally, we suggest to combine both components by setting the time-space-relocation cost to the maximum of both parts, since bike sharing users can wait until a bike is available and relocate to another station during the same time period. The final cost matrix is thus defined as $\Psi_{(i,t_k),(j,t_l)} = \max \{\*C_{ij},\ |t_k - t_l|\}$. 

In \autoref{tab:spacetime}, the N-HiTS model is compared to a linear regression model in their performance of forecasting the bike sharing demand for the next five hours. We evaluate 100 test samples, where each sample comprises all stations for these five time steps, in terms of the MSE (averaging across locations and time), the spatial $\mathrm{W}^{geo}_c$ (average over the five time steps), and the spatio-temporal $\mathrm{W}^{geo}_c$. The results show that the spatial $\mathrm{W}^{geo}_c$ of the linear model is only 18\% larger than the one for N-HiTS, whereas the spatio-temporal error increases by 30\% (67.62 vs 52.14). This result indicates that the N-HiTS model indeed excels in accurate multi-step forecasting due to its hierarchical learning process. More general, this framework allows to quantify and compare temporal and spatial shifts of models proposed for spatio-temporal prediction tasks.

\begin{table}[ht]
\begin{minipage}[b]{0.57\linewidth}
    \centering
    \includegraphics[width=\textwidth]{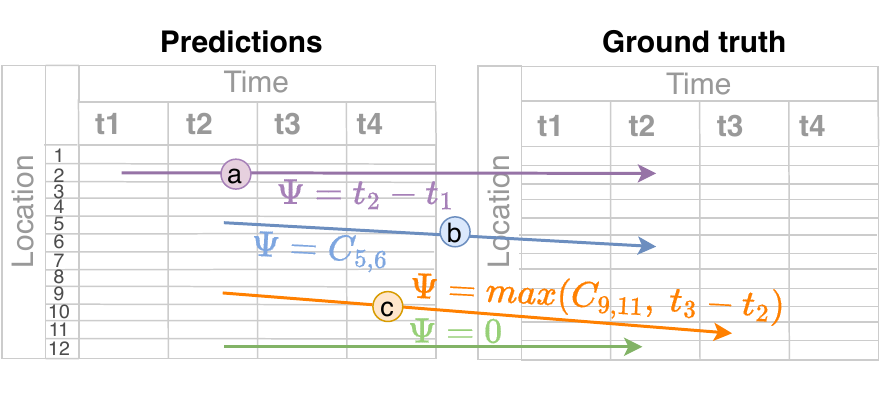}
    \captionof{figure}{OT across space and time. Spatio-temporal cost (c) combines temporal cost (a, e.g. waiting time), with spatial cost (b, e.g. relocation time).}
    \label{fig:spacetime}
\end{minipage}
\hfill
\begin{minipage}[b]{0.4\linewidth}
\centering
\resizebox{\textwidth}{!}{
\begin{tabular}{l|ll}
\toprule
 & N-HiTS & Linear regression \\
\toprule
spatial & 39.05 & 46.07 \\
$\mathrm{W}^{geo}_c$ (mean) & & \\
\midrule
spatio-temporal & 52.14 & 67.62 \\
$\mathrm{W}^{geo}_c$ & & \\
\midrule
MSE & 1.24 & 1.94 \\
\bottomrule
\end{tabular}
}
\vspace{1em}
\caption{Model comparison by spatial and spatio-temporal OT error. N-HiTS particularly improves the spatio-temporal error.}
\label{tab:spacetime}
\end{minipage}
\end{table}

\section{Training with the Sinkhorn divergence}\label{app:sinkhorn}

\subsection{The Sinkhorn algorithm}

As discussed in \S~\ref{subsec:ot-framework}, OT provides a discrepancy measure between two distribution $\mu, \nu$. On the other hand, although solvable as a linear program, Problem~\eqref{eq:discrete-ot} presents some challenges. Firstly, the time complexity of the network simplex algorithm scales as $\mathcal{O}(nm(n+m)\log(n+m))$, where $n$ is the support size of the source distribution $\mu$ and $m$ is that of the target distribution $\nu$. This complexity restricts its applicability to a large number of locations. Secondly, the solution $\*T^\star$ to the OT problem~\ref{eq:discrete-ot} is not necessarily unique. As discussed in \autoref{sec:sinkhorn}, these issues can be circumvented with Entropic OT: 
\begin{equation}
\label{eq:entropic-ot-appendix}
    \mathrm{W}_{c,\varepsilon}(\mu,\nu) = \min_{\*T\in\mathrm{U}(\*p, \*q)} \langle \*T, \*C \rangle\, - \varepsilon H(\*T).
\end{equation}

\begin{wrapfigure}{r}{0.5\textwidth}
\begin{minipage}{0.5\textwidth}
\vspace{-6mm}
\begin{algorithm}[H]
\caption{\textsc{Sink}$(\*p, \*q, \*X, \*Y, \varepsilon, \tau)$.}
\label{algo:sinkhorn}
\begin{algorithmic}[1]
    \State{$\*x_1,\dots,\*x_n=\*X, \quad \*y_1, \dots, \*y_m= \*Y$}
    \State{$(\*f, \*g) \gets (\mathrm{0}_n, \mathrm{0}_m)$}
    \State{$\*{C}\leftarrow [c(\*x_i,\*y_j)]_{ij}$}
    \While{$\|\exp\left(\tfrac{\mathbf{C}-\*f\oplus\*g}{\varepsilon}\right)\mathbf{1}_m-\*p\|_1<\tau$}
        \State $\*f \leftarrow \varepsilon\log \*p - \min_\varepsilon(\mathbf{C}-\*f\oplus\*g) + \*f$
        \State $\*g \leftarrow \varepsilon\log \*q - \min_\varepsilon(\mathbf{C}^\top-\*g\oplus\*f) + \*g$
    \EndWhile
    \State{{\bfseries return} {$\*f, \*g, \*T=\exp\left((\mathbf{C}-\*f\oplus\*g)/\varepsilon\right)$ \label{lst:line:coupling}}}
\end{algorithmic}
\end{algorithm}
\end{minipage}
\end{wrapfigure}

Here, $\varepsilon$ controls the regularization strength. For $\varepsilon \rightarrow 0$, one recovers the standard Wasserstein distance, namely $\mathrm{W}_{c, \varepsilon}(\mu, \nu) \to \mathrm{W}_{c}(\mu, \nu)$. The Entropic OT Prob.~\ref{eq:entropic-ot} admits a dual formulation, which takes the form of an unconstrained, $\varepsilon$-strongly concave program
\begin{align}
\label{eq:dual-entropic-ot}
    \mathrm{W}_{c,\varepsilon}(\mu,\nu) = 
    & \max_{\*f\in\mathbb{R}^n,\*g\in\mathbb{R}^m} \langle \*f, \*p \rangle\ + \langle \*g, \*q \rangle \notag\\
    & - \varepsilon \langle e^{\*f/\varepsilon},\mathbf{K}e^{\*g/\varepsilon}\rangle\,,
\end{align}
where $\mathbf{K} = [\exp(-c(\*x_i, \*y_j)/\varepsilon)]_{1\leq i,j\leq n,m} \in \mathbb{R}^{n\times m}_+$ and $e^\*f, e^\*g$ denotes the element-wise exponential of the vectors $\*f,\*g$. By strong concavity, the optimal $\*f^\star,\*g^\star$ exist and are unique.

\citeauthor{Sinkhorn64}'s algorithm provides an iterative approach for finding $(\*f^\star, \*g^\star)$, which we summarize in Algorithm~\ref{algo:sinkhorn}. For a matrix $\*A = [\*A_{ij}]_{1\leq i,j\leq n,m}$, we define the (rowise) $\varepsilon$-soft-min operator as: $\min_\varepsilon(\mathbf{A}) \coloneqq [-\varepsilon \log\left( \mathbf{1}^\top e^{-\*A_{i,\cdot}/\varepsilon}\right)]_{1\leq i\leq n}$, and $\oplus$ denotes the tensor sum of two vectors, i.e., $\*f \oplus \*g := [\*f_i + \*g_j]_{1\leq i,j\leq n,m}$. Solving the dual Entropic OT Prob.~\eqref{eq:entropic-ot} also provides a valid coupling through the primal-dual relationship: $\*T^\star_\varepsilon = \exp\left((\mathbf{C}-\*f^\star\oplus\*g^\star)/\varepsilon\right)$. Since \citeauthor{Sinkhorn64}'s algorithm essentially alternates between matrix-vector multiplications, its computational complexity scales as $\mathcal{O}(nm)$. Similarly, the memory complexity is also $\mathcal{O}(nm)$, as the cost matrix $\*C$ must be stored.

By \citeauthor{danskin2012theory}'s Theorem, the uniqueness of $\*f^\star$ and $\*g^\star$ guarantees the differentiability of $\mathrm{W}_{c,\varepsilon}(\mu,\nu)$ with respect to its inputs. Moreover, we treat $\*f^\star$ and $\*g^\star$ as constants during differentiation. As a result, there is no need to back-propagate through \citeauthor{Sinkhorn64}'s algorithm. Formally, for any input $\blacksquare$, that can be a location $\*x_i$, $\*y_j$, or a weight vector $\*p,\*q$, one has:
\begin{equation}
    \nabla_\blacksquare \mathrm{W}_{c,\varepsilon}(\mu,\nu) = \nabla_\blacksquare \langle \*f^\star, \*p \rangle\ + \nabla_\blacksquare \langle \*g^\star, \*q \rangle 
    - \varepsilon \nabla_\blacksquare \langle e^{\*f^/\varepsilon},\mathbf{K}e^{\*g^\star/\varepsilon}\rangle\,,
\end{equation}
where both $\*f^\star$ and $\*g^\star$ are treated as constants with respect to\ $\blacksquare$. For instance, the gradients with respect to\ the weight vectors $\*p,\*q$, which are used in our method, are given by:
\begin{equation}
\begin{split}
    & \nabla_\*p \mathrm{W}_{c,\varepsilon}(\mu,\nu) = \nabla_\*p \langle \*f^\star, \*p \rangle = \*f^\star \\
    & \nabla_\*q \mathrm{W}_{c,\varepsilon}(\mu,\nu) = \nabla_\*q \langle \*g^\star, \*q \rangle = \*g^\star
\end{split}
\end{equation}

\subsection{The Sinkhorn divergence}

We recall that when \(c(\mathbf{x}, \mathbf{y}) = \|\mathbf{x} - \mathbf{y}\|_2\) or \(c(\mathbf{x}, \mathbf{y}) = \|\mathbf{x} - \mathbf{y}\|_2^2\), \(\mathrm{W}_{c}(\mu, \nu) \geq 0\) with equality if and only if \(\mu = \nu\). This property is central to our approach, as it justifies the use of \(\mathrm{W}_c\) for comparing distributions. However, this property does not hold when entropy regularization is introduced, as \(\mathrm{W}_{c,\varepsilon}(\mu, \nu)\) can become negative, and in general, \(\mathrm{W}_{c,\varepsilon}(\mu, \mu) \ne 0\).
As a result, this introduces a bias that complicates the use of \(\mathrm{W}_{c,\varepsilon}\) as a loss function.

In light of this phenomenon, several works~\citep{genevay2018learning,feydy2018interpolating,pooladian2022debiaserbewarepitfallscentering} have proposed centering the Entropic OT objective, thereby defining the Sinkhorn divergence as follows:
\begin{equation}
    \mathrm{S}_{c, \varepsilon}(\mu, \nu) = \mathrm{W}_{c,\varepsilon}(\mu, \nu) - \tfrac{1}{2}(\mathrm{W}_{c,\varepsilon}(\mu, \mu) + \mathrm{W}_{c,\varepsilon}(\nu, \nu))
\end{equation}
Centering the Entropic OT objective is akin to debiasing it. For example, when \(c(\mathbf{x}, \mathbf{y}) = \|\mathbf{x} - \mathbf{y}\|_2\) or \(c(\mathbf{x}, \mathbf{y}) = \|\mathbf{x} - \mathbf{y}\|_2^2\), the following holds~\citep[Theorem 1]{feydy2018interpolating}
\begin{equation}
    \mathrm{S}_{c, \varepsilon}(\mu, \nu) \geq 0 \textrm{  with equality if and only if  } \mu = \nu.
\end{equation}

Additionally, the two corrective terms–\(\mathrm{W}_{c,\varepsilon}(\mu, \mu)\) and \(\mathrm{W}_{c,\varepsilon}(\nu, \nu)\)–can be computed even more efficiently than by directly using the Sinkhorn algorithm~\ref{algo:sinkhorn}. As noted by \citet{feydy2018interpolating}, the key observation is that when \(\mu = \nu\), the dual Prob.~\eqref{eq:dual-entropic-ot} reduces to a concave maximization problem with respect to a single variable. This can be solved by iterating a well-conditioned fixed-point update, which typically converges to the desired precision within three iterations.

\section{Computational complexity}\label{app:runtime}

\begin{figure}[htb]
    \centering
    \includegraphics[width=0.6\linewidth]{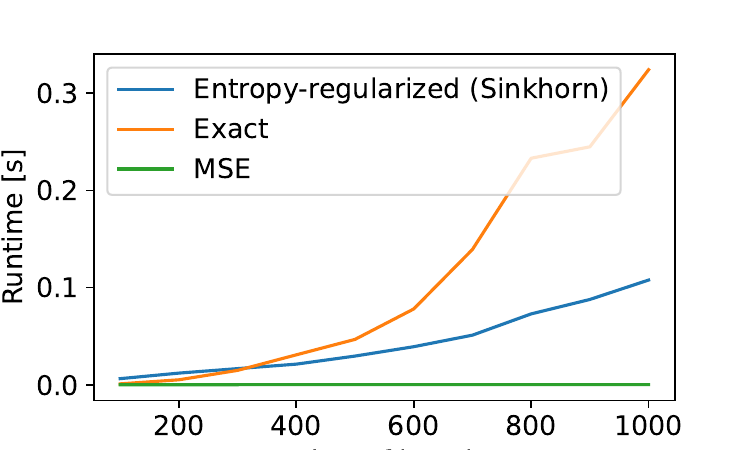}
    \caption{Computational complexity of exact and entropy-regularized OT}
    \label{fig:runtime}
\end{figure}

We evaluated the computational complexity of exact OT (via a linear program) and entropy-regularized OT (Sinkhorn loss) on synthetic data, varying the number of locations from 100 to 1000. For each setup, observations and predictions were sampled randomly 10 times. \autoref{fig:runtime} shows that the Sinkhorn loss runtime scales approximately linearly with the number of locations, whereas the exact computation shows a steeper increase. In our real-world experiments, the largest dataset (bike sharing) had 458 locations. The Sinkhorn loss remains feasible even with up to 1000 locations. During training (on CPU), the Sinkhorn loss is approximately 40 times slower than the standard MSE loss, requiring 13s per iteration compared to 0.33s for MSE. This difference is expected due to the highly optimized MSE implementation in \texttt{torch} compared to the iterative Sinkhorn algorithm. While this increased runtime is a drawback, the Sinkhorn loss offers significant benefits for capturing spatial relationships.

\end{document}